\documentclass{article}

\usepackage{arxiv}

\usepackage[utf8]{inputenc}
\usepackage[T1]{fontenc}
\usepackage[round]{natbib}
\usepackage{hyperref}
\hypersetup{colorlinks=true, linkcolor=blue!60!black, citecolor=blue!60!black, urlcolor=blue!60!black}
\usepackage{url}
\usepackage{booktabs}
\usepackage{amsfonts}
\usepackage{amsmath}
\usepackage{amssymb}
\usepackage{nicefrac}
\usepackage{microtype}
\usepackage{xcolor}
\usepackage{graphicx}
\usepackage{multirow}
\usepackage{enumitem}
\usepackage{amsthm}
\newtheorem{proposition}{Proposition}
\newtheorem{theorem}{Theorem}
\theoremstyle{definition}
\newtheorem{definition}{Definition}

\newcommand{\best}[1]{\textbf{#1}}
\newcommand{\tie}[1]{\emph{#1}}

\newcommand{\ourcf}{\textsc{Ours}}
\newcommand{\lcfnet}{LCF-Net}
\newcommand{\Atilde}{\widetilde{A}}

\title{Closed-Form Node Classification with Exact Graph Unlearning}

\author{%
  Aditya Gaur\thanks{Corresponding author.}\\
  Machine Learning Lab\\
  IIIT Hyderabad\\
  \texttt{aditya.gaur@students.iiit.ac.in}
  \And
  Charu Sharma\\
  Machine Learning Lab\\
  IIIT Hyderabad\\
  \texttt{charu.sharma@iiit.ac.in}
}

\date{}

\begin{document}

\maketitle

\begin{abstract}
Graph neural networks for node classification are typically trained by gradient descent over hundreds or thousands of epochs. Recent work has shown that, when properly tuned, classic GCN/SAGE/GAT architectures can match graph transformers on many node-classification benchmarks. We ask a complementary question: how much of this performance can be recovered by deterministic closed-form solvers, and what guarantees does this enable?

We introduce a routed closed-form framework selected by adjusted homophily. For assortative graphs, we use SGC-style propagation followed by Ridge regression; for heterophilous graphs, we introduce \lcfnet, a layer-wise closed-form graph feature-refinement network whose per-layer Ridge solves are capped by a Gaussian kernel-Ridge head. Across 14 benchmarks, including ogbn-arxiv and ogbn-proteins, our closed-form predictors match or beat the best vanilla 2-layer GCN/SAGE/GAT on 9 of 9 measured datasets, tie tuned deep recipes within one standard deviation on 9 of 12 small benchmarks, and exceed the OGB-leaderboard plain GCN on both large graphs. The remaining heterophilous gap closely tracks the gain from vanilla 2-layer to deep SAGE, suggesting that the residual difference is primarily architectural.

Because our predictors are explicit solutions of deterministic linear systems, modified graph inputs can be re-solved to obtain retrain-equivalent parameters. We formalize exact graph-object unlearning for label, feature, edge, node, and subgraph modifications and prove a K-hop locality theorem for Ridge components (Pipeline~A and the first \lcfnet\ layer); deeper \lcfnet\ layers and the Gaussian KRR head require a closed-form full re-solve, both byte-identical to retraining. We verify exactness across 109 configurations. On ogbn-arxiv, localized updates give $21$--$45\times$ speedups over full re-solving and $\approx\!10^{6}\times$ speedups over gradient retraining-from-scratch in the headline best case. Structural-inversion experiments further quantify the privacy floor of exact retraining and the additional leakage of approximate graph-unlearning methods.
\end{abstract}

\section{Introduction}
\label{sec:intro}

Graph neural networks (GNNs) are the dominant approach to node classification, and they are trained with gradient descent for hundreds to thousands of epochs. \citet{luo2024classic} recently showed that this gradient training, when combined with modern tricks (BatchNorm, residual connections, depth sweeps, dropout), lets classic GCN/SAGE/GAT architectures match or exceed graph transformers on 17 of 18 node-classification benchmarks. Following Luo's rigor-promoting precedent, we ask a complementary question: if propagation $\widetilde A^k X$ already encodes most of the class signal, can a \emph{closed-form solver} recover the same predictor without gradient descent---and obtain machine-learning guarantees that gradient-trained GNNs cannot, namely \emph{exact unlearning}?

We propose two closed-form pipelines selected by adjusted homophily~\citep{platonov2022characterizing} on the 12 small benchmarks plus ogbn-arxiv (ogbn-proteins is hand-assigned to Pipeline~B; multi-label, no scalar $h_{\mathrm{adj}}$). Pipeline~A (SGC~\citep{wu2019sgc} + Ridge + Correct-and-Smooth~\citep{huang2021combining}) handles assortative graphs. Pipeline~B is our novel \emph{Layer-wise Closed-Form Deep Network} (\lcfnet): each gradient-trained $W_k$ in a deep GCN is replaced by a per-layer Ridge solve against the training labels, capped by a Gaussian kernel-Ridge head. Crucially, we use \emph{adjusted homophily} $h_{\mathrm{adj}}$~\citep{platonov2022characterizing}---the field-standard assortativity-style measure that, unlike edge homophily, is class-count invariant---which routes Questions ($h_{\mathrm{adj}}\!\approx\!0.02$) into the heterophilous pipeline despite its $h_{\mathrm{edge}}\!=\!0.84$.

\textbf{Empirical headline.} Across 14 benchmarks at matched shallow depth, closed-form ties tuned deep recipes within one standard deviation on $9$ of $12$ small benchmarks (Table~\ref{tab:main}); on Minesweeper and Roman-empire the residual $7$--$8$\,pp gap tracks the depth gain from vanilla 2-layer SAGE to deep SAGE, attributing it to architectural depth, not training. At OGB scale, TICR-Multi (Transductive Iterative Closed-form Refinement) reaches $71.91 \pm 0.09\%$ on ogbn-arxiv and LP-Ridge reaches $74.87$ ROC--AUC on ogbn-proteins, both exceeding the OGB-leaderboard plain GCN.

\textbf{Exact graph-object unlearning.} Because every component is a deterministic function of the training data, removing a record and re-solving yields weights byte-identical to retraining from scratch. We formalize this as exact unlearning~\citep{bourtoule2021machine}, prove it holds across all five graph-object modifications (label, feature, edge, node, subgraph; Proposition~\ref{prop:five-types}), and prove a K-hop locality theorem for Ridge components (Theorem~\ref{thm:khop-locality}): for Pipeline~A and the first \lcfnet\ layer, the exact Ridge refresh runs in $O(|N_L(S)| D^2)$ from the $L$-hop neighborhood, with weights and predictions byte-identical to retraining. Deeper \lcfnet\ layers and the Gaussian KRR head fall back to a closed-form full re-solve, also byte-identical to retraining. We verify byte-identical weights and predictions across $109$ configurations with MIA-AUC at chance (Appendix~\ref{app:unlearning-full}), and observe $21$--$45\times$ speedup on ogbn-arxiv from K-hop locality vs.\ a full re-solve ($\approx\!10^{6}\times$ vs.\ gradient retrain-from-scratch in the headline best case).

\paragraph{Contributions.}
\begin{itemize}[leftmargin=2em,topsep=0pt,itemsep=2pt]
  \item[\textbf{(C1)}] \textbf{\lcfnet, a layer-wise closed-form deep graph network.} Each gradient-trained $W_k$ in a deep GCN is replaced by a per-layer Ridge solve against the training labels, capped by a Gaussian kernel-Ridge head; depth performs label-supervised iterative refinement without any gradient descent. \lcfnet\ closes $49\%$ of the mean gap to tuned deep GCN on the four heterophilous datasets where prior closed-form approaches plateau (per-dataset closed-gap fractions $73 / 66 / 42 / 14\%$; $+3.19$\,pp mean over prior-best closed-form).
  \item[\textbf{(C2)}] \textbf{14-benchmark, three-architecture evaluation.} We locally reproduce all 33 tuned recipes of~\citet{luo2024classic} (11 datasets $\times$ GCN/SAGE/GAT) within $\pm 1\sigma$ of published numbers, add 9 vanilla 2-layer fairness peers, and extend to two OGB benchmarks. Closed-form matches or beats the best vanilla 2-layer architecture on \textbf{$9$ of $9$} measured datasets and exceeds the OGB-leaderboard plain GCN on ogbn-arxiv (+$0.17$\,pp) and ogbn-proteins (+$2.36$\,pp).
  \item[\textbf{(C3)}] \textbf{Architectural-depth attribution.} The residual gap to deep recipes on Minesweeper / Roman-empire is architectural depth, not training-vs-training-free: vanilla 2-layer SAGE has the same $+7.3$ to $+7.5$\,pp gap to deep SAGE that closed-form has to deep SAGE.
  \item[\textbf{(C4)}] \textbf{Exact unlearning for five graph-object modifications.} Definition~\ref{def:exact-unlearning} (exact unlearning) and Proposition~\ref{prop:five-types} formalize byte-identical retrain-equivalence under any of label, feature, edge, node, or subgraph deletion. Empirical verification across $109$ configurations: byte-identical weights and probabilities; argmax agreement $107/109$; MIA-AUC at chance on $108/109$ rows.
  \item[\textbf{(C5)}] \textbf{K-hop locality theorem for Ridge components.} Theorem~\ref{thm:khop-locality}: for Pipeline~A and the first \lcfnet\ Ridge solve, the exact refresh runs in $O(|N_L(S)| D^2)$ from the $L$-hop neighborhood, with weights and predictions byte-identical to retraining. Deeper \lcfnet\ layers and the Gaussian KRR head require a closed-form full re-solve, also byte-identical to retraining. Empirically: $21$--$45\times$ wall-clock speedup over full re-solve on ogbn-arxiv ($\approx\!10^{6}\times$ over gradient retrain-from-scratch in the headline best case).
  \item[\textbf{(C6)}] \textbf{Empirical privacy-floor measurement under structural attack.} Pipeline~A's byte-identity to retrain-from-scratch supplies the empirical privacy-floor reference data point that~\citet{li2025trendattack} (TrendAttack) Table~1 omits. Across 12 attack configurations, we quantify approximate-unlearning over-leakage at $0.11$--$0.18$ AUC on Cora and CiteSeer under TrendAttack-MIA---the first such quantification in the GNN unlearning literature.
\end{itemize}

\paragraph{Paper organization.} \S\ref{sec:related} situates the work; \S\ref{sec:method} defines the two pipelines and Theorem~\ref{thm:khop-locality}; \S\ref{sec:setup-exp}--\ref{sec:results} present empirical results, OGB-scale evidence, exact-unlearning verification, and TrendAttack analysis; \S\ref{sec:limitations} discusses limitations; \S\ref{sec:conclusion} concludes.

\section{Related work}
\label{sec:related}

\paragraph{Closed-form, shallow, and routed GNNs.} \citet{wu2019sgc} introduced SGC; \citet{huang2021combining} proposed C\&S; APPNP~\citep{klicpera2019appnp} and GPR-GNN~\citep{chien2021gprgnn} generalize SGC with personalized PageRank. These plateau on heterophily, motivating heterophily-specific architectures (FAGCN~\citep{bo2021fagcn}, H$^2$GCN~\citep{zhu2020h2gcn}, ACM-GCN~\citep{luan2022acmgcn}) and routed/mixture variants (Mowst~\citep{zeng2024mowst}, Node-MoE~\citep{wang2024nodemoe}, GraphAny~\citep{zhao2025graphany}). \citet{luo2024classic} show tuned classic GCN/SAGE/GAT match or beat specialized heterophily models and graph transformers (GraphGPS~\citep{rampasek2022graphgps}, SGFormer~\citep{wu2023sgformer}, Polynormer~\citep{deng2024polynormer}); we use Luo's recipes as baselines. The adjusted-homophily measure $h_{\mathrm{adj}}$~\citep{platonov2023critical,platonov2022characterizing} drives our routing rule. Distinct from random-feature stacks (ELM~\citep{huang2006elm}, reservoir/ESN~\citep{jaeger2001echo}), \lcfnet's per-layer $W_k$ is the Ridge solution \emph{against training labels}, not a random matrix; depth performs label-supervised refinement (the frozen-random ablation underperforms \lcfnet\ by $5$--$11$\,pp on heterophilous datasets, Appendix~\ref{app:failed}). Distinct from training-free kernel methods (GNTK~\citep{du2019gntk}, RFF~\citep{rahimi2007random}), \lcfnet\ uses practical finite-width per-layer Ridge rather than an infinite-width kernel, avoiding the $O(n^2 L)$ memory cost. Pipeline~A is pure Ridge on top of SGC propagation---fully training-free including the base predictor, unlike C\&S's gradient-trained MLP base. Our hard router (Pipeline~A vs.\ Pipeline~B by $h_{\mathrm{adj}}$) is a practical lower bound; a learned per-node soft-mixture (Appendix~\ref{app:soft-mixture}) recovers it within noise.

\paragraph{Graph unlearning.} \emph{Approximate} methods---GraphEraser~\citep{chen2022graph}, GNNDelete~\citep{cheng2023gnndelete}, GIF~\citep{wu2023gif}, Certified Edge Unlearning~\citep{wu2023ceu}, MEGU~\citep{li2024megu}, IDEA~\citep{dong2024idea}, CGU~\citep{chien2023cgu}---modify the trained GNN via shard-retraining, layer-wise edits, or convex-loss perturbation, leaving a detectable MIA fingerprint ($\geq\!0.65$). \emph{Exact} unlearning~\citep{bourtoule2021machine,sekhari2021remember} requires byte-identity to retrain. The closest precedent is GraphEditor~\citep{cong2023grapheditor}, which studies exact graph representation editing and unlearning for linear GNNs (architecturally restricted to a single linear layer). Our Theorem~\ref{thm:khop-locality} generalizes to all five graph-object modifications (label/feature/edge/node/subgraph) for the Ridge components of both pipelines (Pipeline~A and the first \lcfnet\ layer), with $K$-hop locality matched to GNN depth; deeper \lcfnet\ layers and the Gaussian KRR head fall back to a closed-form full re-solve, byte-identical to retraining. ScaleGUN~\citep{zhang2025scalegun} achieves locality via shard-cached propagation but is approximate within shards. \citet{li2025trendattack} (TrendAttack) shows all current unlearning methods are vulnerable to structural-inversion attacks; \S\ref{sec:limitations} discusses this orthogonal frontier.

\section{Method}
\label{sec:method}

\begin{figure}[t]
  \centering
  \includegraphics[width=\textwidth,trim=0 10 0 10,clip]{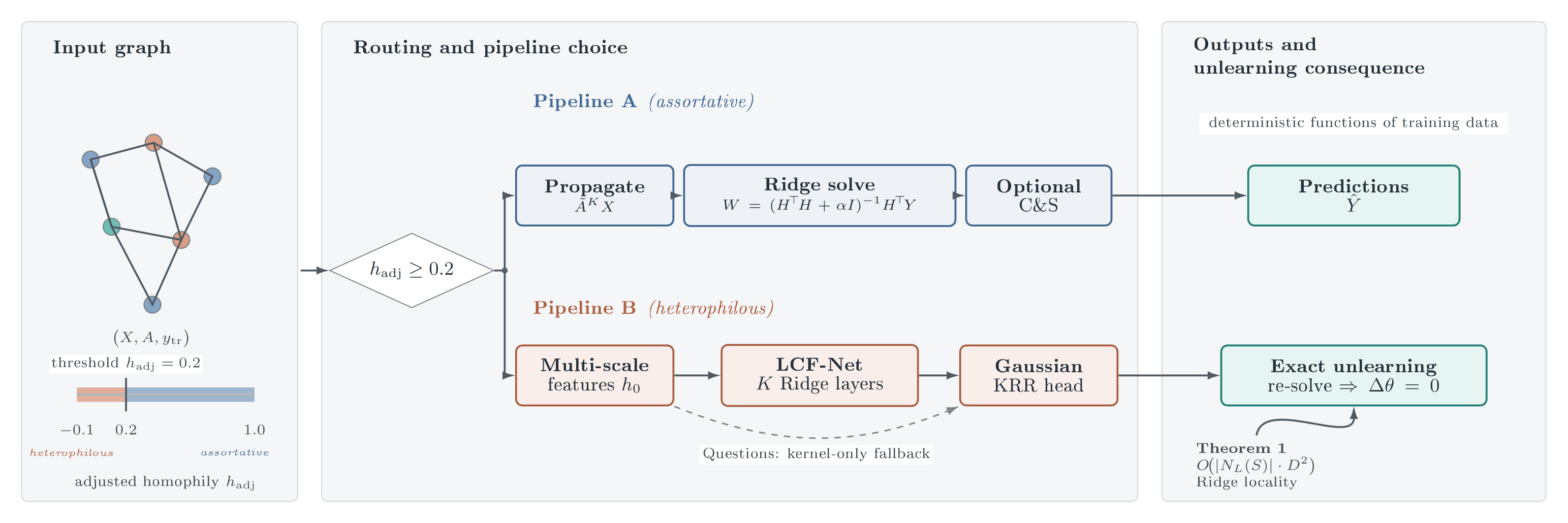}
  \caption{\textbf{Method overview.} Routing on $h_{\mathrm{adj}}$ at $\tau{=}0.2$ selects Pipeline~A (assortative; SGC+Ridge+C\&S) or Pipeline~B (heterophilous; multi-scale + \lcfnet + Gaussian KRR). Theorem~\ref{thm:khop-locality}: exact unlearning refresh in $O(|N_L(S)| D^2)$ for Ridge components, byte-identical to retrain.}
  \label{fig:method}
\end{figure}

We propose a two-pipeline closed-form approach selected by adjusted homophily (Figure~\ref{fig:method}); ogbn-proteins is hand-assigned to Pipeline~B (multi-label, no scalar $h_{\mathrm{adj}}$). Pipeline~A (\S\ref{sec:pipeline-a}) targets assortative graphs ($h_{\mathrm{adj}} \geq 0.2$): SGC-style propagation followed by Ridge regression and optional C\&S. Pipeline~B (\S\ref{sec:pipeline-b}) targets heterophilous graphs ($h_{\mathrm{adj}} < 0.2$): our novel \emph{Layer-wise Closed-Form Deep Network} (\lcfnet), stacking per-layer Ridge solves capped by a Gaussian kernel-Ridge head; val-selection chooses \lcfnet+KRR or KRR-only. The threshold $\tau{=}0.2$ is fixed a priori; routing is invariant for $\tau \in [0.16, 0.4]$ (Appendix~\ref{app:routing}). Both pipelines are deterministic, giving the exact-unlearning guarantee of \S\ref{sec:unlearning-method}.

\paragraph{Setup.} Transductive node classification on graph $G=(V,E)$ with $n=|V|$, symmetric adjacency $A$, self-loop-augmented normalized propagation $\Atilde = \bar{D}^{-1/2}(A+I)\bar{D}^{-1/2}$, features $X\in\mathbb{R}^{n\times d}$, train set $V_{\mathrm{tr}}$ ($|V_{\mathrm{tr}}|=n_{\mathrm{tr}}$), one-hot labels $Y_{\mathrm{tr}}$, and row-$L_2$-normalized features $\hat{X}$. The \emph{adjusted homophily} of~\citet{platonov2022characterizing} corrects edge-homophily for class-count imbalance and drives our routing rule (full formula and per-dataset values in Table~\ref{tab:datasets} and Appendix~\ref{app:hadj}); on Questions $h_{\mathrm{adj}}{=}0.02$ (heterophilous, despite $h_{\mathrm{edge}}{=}0.84$), and on Cora $h_{\mathrm{adj}}{=}0.77$.

\subsection{Pipeline A: SGC + Ridge + Correct-and-Smooth (\texorpdfstring{$h_{\mathrm{adj}} \geq 0.2$}{h\_adj >= 0.2})}
\label{sec:pipeline-a}

For assortative graphs ($h_{\mathrm{adj}} \geq 0.2$), neighbors are more likely than chance to share class labels, so $k$-hop smoothing $\Atilde^k X$ amplifies the class signal. Pipeline~A exploits this with a closed-form linear model, following and slightly extending SGC \citep{wu2019sgc} and C\&S \citep{huang2021combining}:
\begin{align}
  H &= \Atilde^K \cdot X_{\mathrm{src}}, \qquad X_{\mathrm{src}} \in \{X, \hat{X}\}, \quad K \in \{1, \ldots, 8\}
  \label{eq:sgc-feat} \\
  W &= \bigl(H_{\mathrm{tr}}^\top H_{\mathrm{tr}} + \alpha I\bigr)^{-1} H_{\mathrm{tr}}^\top Y^{(\varepsilon)}_{\mathrm{tr}}
  \label{eq:ridge}
\end{align}
with label smoothing $Y^{(\varepsilon)} = (1-\varepsilon) Y + \varepsilon / C$ ($C$ the number of classes), and Ridge regularizer $\alpha$. Raw predictions are $\hat{Y} = H W$. Optionally, we apply Correct-and-Smooth \citep{huang2021combining} as a post-hoc pass. The tuple $(X_{\mathrm{src}}, K, \alpha, \varepsilon, \mathrm{C\&S~params})$ is selected on the validation set. For two of the seven datasets (CiteSeer, Amazon-Photo) the winning variant uses a richer feature map---multi-hop concatenation with per-group Tikhonov regularization, or Gaussian random Fourier features \citep{rahimi2007random}---detailed in Appendix~\ref{app:pipeline-a-variants}.

Pipeline~A contains no learned nonlinearity and no gradient-trained base predictor; Ridge is a single linear-system solve. This distinguishes it from C\&S~\citep{huang2021combining}, whose standard recipe uses a gradient-trained MLP base. SGC+Ridge is mathematically equivalent to kernel Ridge with the graph-diffusion kernel $K=\Phi\Phi^\top$, $\Phi=\Atilde^K\hat{X}$~\citep{wu2019sgc}; the training-free gain is wall-clock, not function class.

\subsection{Pipeline B: \lcfnet\ + Gaussian kernel-Ridge head (\texorpdfstring{$h_{\mathrm{adj}} < 0.2$}{h\_adj < 0.2})}
\label{sec:pipeline-b}

\begin{figure}[!ht]
  \centering
  \includegraphics[width=\linewidth,trim=0 10 0 10,clip]{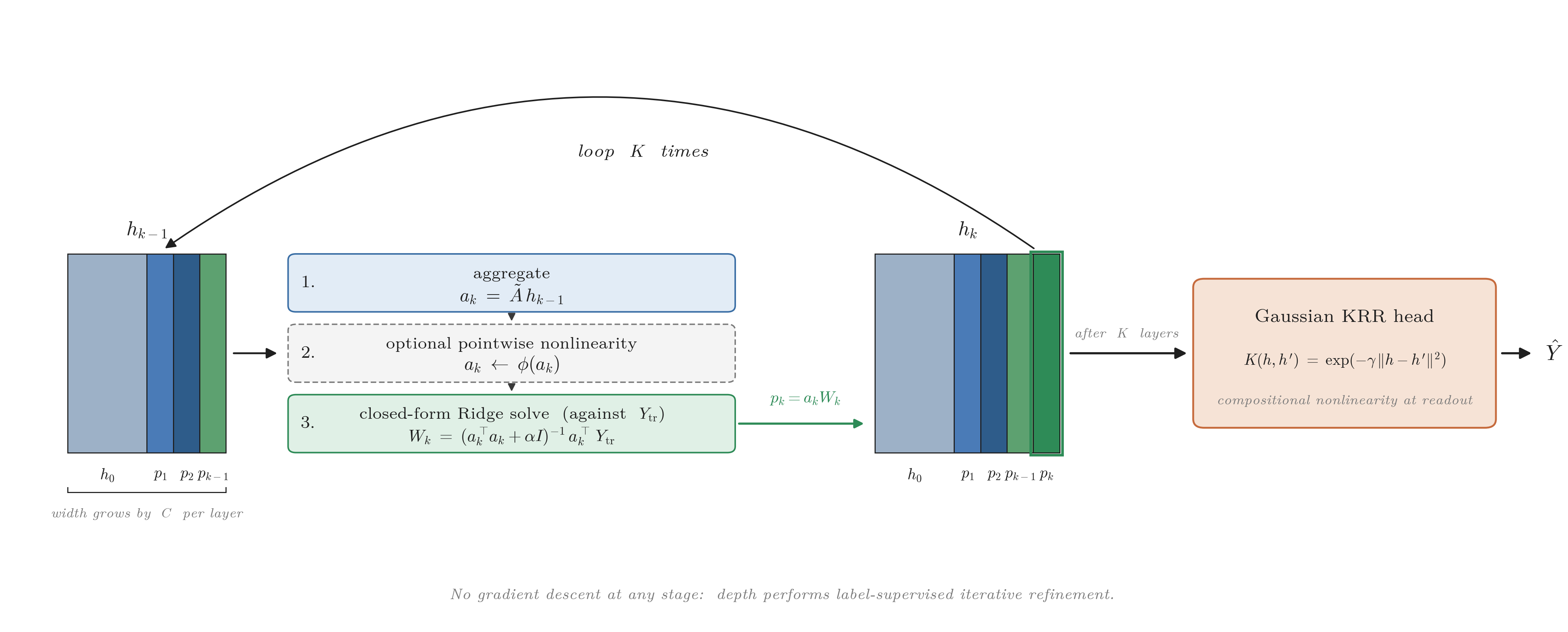}
  \caption{\lcfnet\ layer structure. Each iteration aggregates with $\Atilde$, applies an optional pointwise nonlinearity, solves a closed-form Ridge problem against training labels for $W_k$, and \emph{appends} (residual concatenation) the per-layer prediction $p_k$ to the running representation $h_{k-1}$. Width grows by $C$ columns per layer. After $K$ rounds, a Gaussian kernel-Ridge head provides compositional nonlinearity at readout. No gradient descent at any stage; depth performs label-supervised iterative refinement, distinct from ELM-style random projection.}
  \label{fig:lcfnet}
\end{figure}

For heterophilous graphs, simple smoothing blurs the class signal and linear models saturate. A deep trained GCN addresses this with $K$ layers, each contributing a learned projection $W_k$ that extracts class-discriminative features at depth~$k$. We ask: can we achieve this \emph{without gradient descent}? Figure~\ref{fig:lcfnet} (below) visualizes the per-layer structure of \lcfnet; the full equations follow.

\paragraph{Layer-wise Closed-Form Deep Network (\lcfnet).} Yes---replace each gradient-trained $W_k$ with a \emph{closed-form Ridge solve} against the training labels. We motivate this as a closed-form analogue to greedy layer-wise feature learning \citep{bengio2007greedy}: each layer maps its aggregated input to a label-aligned subspace via a single linear-system solve. We do \emph{not} claim this is a theoretical equivalence to end-to-end-trained $W_k$ (SGD optimizes a different objective), only an empirically motivated heuristic. Unlike ELM-style random-projection stacks, each per-layer $W_k$ is the Ridge solution against training labels, so depth performs label-supervised iterative refinement rather than random expansion. We start from a rich feature base $h_0$ (multi-scale propagation, variance moments, feature-similarity-weighted aggregation; Appendix~\ref{app:lcfnet-base}), then iterate for $K$ rounds:
\begin{align}
  a_k &= \Atilde \cdot h_{k-1} \label{eq:lcf-agg} \\
  a_k &\leftarrow \phi(a_k) \qquad \text{(optional pointwise nonlinearity: none, tanh, or elu)} \label{eq:lcf-nonlin}\\
  W_k &= \bigl(a_k[\mathrm{tr}]^\top a_k[\mathrm{tr}] + \lambda I\bigr)^{-1} a_k[\mathrm{tr}]^\top Y_{\mathrm{tr}}
  \label{eq:lcf-ridge} \\
  p_k &= a_k \cdot W_k \in \mathbb{R}^{n \times C} \label{eq:lcf-pred}\\
  h_k &= [\,h_{k-1} \parallel p_k\,]  \label{eq:lcf-residual}
\end{align}
where $a_k[\mathrm{tr}]$ denotes the sub-matrix restricted to training rows. Equation~\eqref{eq:lcf-residual} implements a residual via column concatenation: the representation $h_{k-1}$ is preserved, and the soft layer-$k$ predictions $p_k$ are appended. After $K$ rounds, we cap the network with exact Gaussian kernel Ridge regression on the final representation $h_K$:
\begin{equation}
  K_{ij} = \exp\!\Bigl(-\|h_K^{(i)} - h_K^{(j)}\|^2 / (2\sigma^2)\Bigr), \quad
  \boldsymbol{\alpha}_{\mathrm{dual}} = (K_{\mathrm{tr}} + \lambda' I)^{-1} Y_{\mathrm{tr}}, \quad
  \hat{Y} = K_{\cdot,\mathrm{tr}} \boldsymbol{\alpha}_{\mathrm{dual}}.
  \label{eq:krr}
\end{equation}
The final head contributes the compositional nonlinearity that the per-layer Ridge solves alone cannot. All hyperparameters ($K$, $\phi$, $\lambda$, $\sigma$, $\lambda'$) are selected on the validation set.

Eqs.~\eqref{eq:lcf-ridge} and~\eqref{eq:krr} are single linear-system solves (no gradient descent); the output is deterministic up to floating-point non-associativity. After $K$ rounds, $h_K \in \mathbb{R}^{n \times (d_0 + KC)}$ concatenates per-depth class-discriminative projections (analogous to Jumping Knowledge Networks~\citep{xu2018jknet}). Per-layer cost is $O(D_k^2 n_{\mathrm{tr}})$, where $D_k = d_0 + (k{-}1)C$ is the running width at layer $k$; the KRR head is $O(n_{\mathrm{tr}}^3)$ with row-chunked prediction (Appendix~\ref{app:compute}).

\paragraph{When \lcfnet\ helps.} Routing at $\tau{=}0.2$ sends all five Pipeline-B small datasets (Minesweeper, Tolokers, Amazon-Ratings, Roman-empire, Questions; $h_{\mathrm{adj}} \in [-0.05, 0.14]$) to Pipeline~B. Val-selection within Pipeline~B picks \lcfnet+KRR on Minesweeper/Tolokers/Amazon-Ratings/Roman-empire (gains $+1$ to $+6$\,pp; \S\ref{sec:lcfnet-results}) and KRR-only on Questions, where high-degree-node distortion of $h_{\mathrm{adj}}$~\citep{platonov2022characterizing} makes simple smoothing suffice. Sensitivity over $\tau\in\{-0.1, \ldots, 0.5\}$ and learned soft-mixture variant in Appendix~\ref{app:routing}, \ref{app:soft-mixture}.

\subsection{Exact unlearning and K-hop locality}
\label{sec:unlearning-method}

We pursue four distinct notions of unlearning exactness, only the first three of which we contribute:
\begin{itemize}[leftmargin=2em,topsep=2pt,itemsep=2pt]
  \item[\textbf{(i)}] \textbf{Weight exactness.} Byte-identical Ridge weights vs.\ retrain-from-scratch. Formalized by Definition~\ref{def:exact-unlearning} and proved by Proposition~\ref{prop:five-types} across all five graph-object modifications.
  \item[\textbf{(ii)}] \textbf{Prediction agreement.} Byte-identical underlying probabilities; argmax agreement modulo decision-boundary ties ($107/109$ configurations; \S\ref{sec:unlearning-results}).
  \item[\textbf{(iii)}] \textbf{MIA indistinguishability.} Membership-inference AUC at chance vs.\ retrain-from-scratch ($108/109$ configurations at chance; \S\ref{sec:unlearning-results}).
  \item[\textbf{(iv)}] \textbf{Structural inference resistance.} Resistance to attacks reconstructing deleted graph elements from black-box query access~\citep{li2025trendattack}. This is an orthogonal property; no aggregation-based GNN---including ours---provides it. Pipeline~A's byte-identity to retrain-from-scratch (\S\ref{sec:trendattack}) supplies the empirical privacy-floor reference for (iv), and we quantify approximate-method over-leakage at $0.11$--$0.18$ AUC versus this floor under TrendAttack-MIA; we do not contribute (iv).
\end{itemize}
Approximate graph-unlearning methods (GraphEraser, GNNDelete, GIF, CEU, MEGU, IDEA) retain a detectable MIA fingerprint on (iii) (typically $\geq\!0.65$). We pursue (i)--(iii) in the formal sense of~\citet{bourtoule2021machine,sekhari2021remember}.

\begin{definition}[Bourtoule--SISA-distributional exact unlearning]
\label{def:exact-unlearning}
Let $\mathcal{A}$ be a (possibly randomized) learner that maps a training set $\mathcal{D}$ to a parameter $\theta$. An \emph{exact} unlearning procedure $\mathcal{U}$ for forget set $\mathcal{F}\subseteq\mathcal{D}$ is one whose output distribution coincides with retraining: $\mathcal{U}(\mathcal{A}(\mathcal{D}), \mathcal{F}) \stackrel{d}{=} \mathcal{A}(\mathcal{D}\setminus\mathcal{F})$. For a deterministic $\mathcal{A}$, this is byte-identity: $\mathcal{U}(\theta, \mathcal{F}) = \mathcal{A}(\mathcal{D}\setminus\mathcal{F})$ up to floating-point non-associativity.
\end{definition}

\begin{proposition}[Exact unlearning across five graph-object modifications]
\label{prop:five-types}
Let $f_\theta$ denote either Pipeline-A or Pipeline-B with parameters $\theta\in\{W, \boldsymbol{\alpha}_{\mathrm{dual}}\}$ obtained by solving the corresponding linear system. For any forget set $\mathcal{F}$ corresponding to one of (i)~labels, (ii)~node features, (iii)~edges, (iv)~nodes (with incident edges), or (v)~induced subgraphs, the unlearning procedure that re-solves the linear system on the modified inputs $(X', A', Y'_{\mathrm{tr}})$ produces parameters $\theta'$ satisfying Definition~\ref{def:exact-unlearning}.
\end{proposition}
The proof, immediate from the deterministic-function structure of Eqs.~\eqref{eq:ridge} and~\eqref{eq:krr}, is in Appendix~\ref{app:unlearning-proof}. We can do strictly better than blind re-solving:

\begin{theorem}[K-hop locality of Ridge components]
\label{thm:khop-locality}
Let $\mathcal{M}$ be a graph-object modification with affected node set $S \subseteq V$, $L$ the Ridge solve's propagation depth. \textbf{(a)~Containment:} rows of $\widetilde{A}^L X$ whose values change under $\mathcal{M}$ lie in $N_L(S) := \{v: d_G(v,S) \leq L\}$ (Pipeline~A end-to-end; first \lcfnet\ Ridge solve). \textbf{(b)~Local update:} under the Ridge normal equations, a rank-$|N_L(S)|$ Schur-complement / SMW refresh updates the Gram matrix and right-hand side in $O(|N_L(S)| D^2)$; the resulting $W$ and the predictions over the entire graph are byte-identical to retraining from scratch on $\mathcal{D} \setminus \mathcal{F}$. \textbf{(c)~Gaussian KRR head:} pair-dependent and inherits no local update; full re-solve in $O(N_{\mathrm{tr}}^3)$ remains retrain-equivalent.
\end{theorem}

\paragraph{Scope at deeper \lcfnet\ layers.} For $k \geq 2$, $W_k$ is a global $D\!\times\!C$ matrix: any change to a row of $a_k$ produces a new $W_k$ that re-multiplies every $a_k(v)$, so $p_k(v) = a_k(v) W_k$ shifts globally and the cascade propagates. Layers $k\!\geq\!2$ therefore require full re-solve, byte-identical to retraining (Proposition~\ref{prop:five-types}); only the locality-based speedup of (b) does not apply. Theorem~\ref{thm:khop-locality}(a)--(b) generalizes GIF's~\citep{wu2023gif} approximate locality to exact for Pipeline~A and \lcfnet\ layer~1, matching ScaleGUN~\citep{zhang2025scalegun} without a propagation cache. Empirically (\S\ref{sec:unlearning-results}, Table~\ref{tab:unlearning}): byte-identical weights and predictions across 109 configurations, $21$--$45\times$ K-hop speedup vs.\ full re-solve, $\approx\!10^{6}\times$ vs.\ gradient retrain-from-scratch.

\section{Experimental setup}
\label{sec:setup-exp}

\paragraph{Datasets.} We evaluate on the 12 small-graph benchmarks used by \citet{luo2024classic} plus two OGB benchmarks~\citep{hu2020ogb}: ogbn-arxiv ($n{=}169\,k$, year-based split, 40 classes) and ogbn-proteins ($n{=}132\,k$, species-based split, 112 multi-label tasks). For the small-graph suite we exclude Squirrel/Chameleon, whose filtered versions \citet{platonov2023critical} note differ non-trivially across releases. Datasets span $h_{\mathrm{adj}} \in [-0.05, 0.85]$ where defined and $n \in [2.7\text{k}, 169\text{k}]$; see Table~\ref{tab:datasets}. Seven small-graph datasets plus Amazon-Ratings and Roman-empire (multi-class) and ogbn-arxiv use accuracy; the three Platonov binary datasets (Minesweeper, Tolokers, Questions) and ogbn-proteins use ROC--AUC. Splits and metrics match \citet{luo2024classic} for the small-graph suite and the OGB-leaderboard protocol for OGB.

Datasets, splits, and routing assignments are summarized in Table~\ref{tab:datasets}. Pipeline-A receives the eight $h_{\mathrm{adj}} \geq 0.2$ datasets; Pipeline-B receives the six $h_{\mathrm{adj}} < 0.2$ datasets. Within Pipeline~B, val-selection picks KRR-only on Questions and \lcfnet+KRR on the other four heterophilous datasets.

\begin{table}[h]
  \caption{The $14$ node-classification benchmarks. $h_{\mathrm{adj}}$: adjusted homophily~\citep{platonov2022characterizing}. Routing: $h_{\mathrm{adj}} \geq 0.2 \to$ Pipeline~A; $h_{\mathrm{adj}} < 0.2 \to$ Pipeline~B.}
  \label{tab:datasets}
  \centering
  \scriptsize
  \setlength{\tabcolsep}{3pt}
  \begin{tabular}{llrrrrrll}
    \toprule
    Routing & Dataset & $n$ & $d$ & $C$ & $h_{\mathrm{edge}}$ & $h_{\mathrm{adj}}$ & Metric & Split \\
    \midrule
    \multirow{8}{*}{Pipeline A} &
      Cora             & 2{,}708   & 1{,}433 & 7  & 0.81 & 0.77       & acc      & seed-123 class rand \\
    & CiteSeer         & 3{,}327   & 3{,}703 & 6  & 0.74 & 0.67       & acc      & seed-123 class rand \\
    & PubMed           & 19{,}717  & 500     & 3  & 0.80 & 0.69       & acc      & seed-123 class rand \\
    & Amazon-Photo     & 7{,}650   & 745     & 8  & 0.83 & 0.78       & acc      & fixed \texttt{.npz} \\
    & Coauthor-CS      & 18{,}333  & 6{,}805 & 15 & 0.81 & 0.76       & acc      & fixed \texttt{.npz} \\
    & Coauthor-Physics & 34{,}493  & 8{,}415 & 5  & 0.93 & 0.85       & acc      & fixed \texttt{.npz} \\
    & WikiCS           & 11{,}701  & 300     & 10 & 0.66 & 0.57       & acc      & predef.\ 0--4 \\
    \cmidrule{2-9}
    & ogbn-arxiv       & 169{,}343 & 128     & 40 & 0.65 & 0.41       & acc      & year-based \\
    \midrule
    \multirow{6}{*}{Pipeline B} &
      Minesweeper      & 10{,}000  & 7       & 2  & 0.68 & 0.01       & ROC--AUC & predef.\ 0--4 \\
    & Tolokers         & 11{,}758  & 10      & 2  & 0.60 & 0.09       & ROC--AUC & predef.\ 0--4 \\
    & Amazon-Ratings   & 24{,}492  & 300     & 5  & 0.38 & 0.14       & acc      & predef.\ 0--4 \\
    & Roman-empire     & 22{,}662  & 300     & 18 & 0.05 & ${-}0.05$  & acc      & predef.\ 0--4 \\
    & Questions        & 48{,}921  & 301     & 2  & 0.84 & 0.02       & ROC--AUC & predef.\ 0--4 \\
    \cmidrule{2-9}
    & ogbn-proteins    & 132{,}534 & 8       & 112 & ---  & ---       & ROC--AUC & species-based \\
    \bottomrule
  \end{tabular}
\end{table}

\paragraph{Baselines.} We compare against tuned GCN, GraphSAGE, and GAT using the recipes of \citet{luo2024classic}. These use per-dataset hyperparameters (depth, hidden dimension, BatchNorm/LayerNorm, residuals, dropout, learning rate) tuned to achieve published state-of-the-art for shallow-to-deep GNNs. We reproduce all 33 recipes (11 datasets $\times$ 3 architectures; Tolokers has no published recipe, so we adopt a standard heterophilous-configuration (hidden $=128$, 4 layers, BatchNorm $+$ residuals, 2000 epochs) uniformly across the three architectures). Every reproduction falls within one standard deviation of the published number (Appendix~\ref{app:reproductions}). To establish architectural fairness we additionally run \emph{vanilla 2-layer} GCN, SAGE, and GAT (no BatchNorm, LayerNorm, residuals, or pre-linear projection) on the three hetero datasets where the tuned Luo recipe is $\ge 4$ layers (Minesweeper, Amazon-Ratings, Roman-empire).

\paragraph{Hyperparameter selection.} All method hyperparameters (propagation depth $K$, ridge regularizer $\alpha$, kernel bandwidth $\sigma$, label-smoothing $\varepsilon$, Correct-and-Smooth parameters, \lcfnet\ nonlinearity and regularizer) are selected on the validation set. No hyperparameter is tuned per-dataset by hand; the same grid is swept across datasets. See Appendix~\ref{app:hparams}.

\paragraph{Protocol note on variance.} Closed-form methods are deterministic given a fixed split: solving the linear system yields a unique weight matrix. For datasets with a single fixed split (Cora/CiteSeer/PubMed under \citet{luo2024classic}'s \texttt{rand\_split\_class(seed=123)} protocol; Amazon-Photo, Coauthor-CS, Coauthor-Physics) we therefore report a single number without error bars. For datasets with multiple predefined splits (WikiCS, all five heterophilous datasets) we report mean $\pm$ standard deviation across 5 splits. Luo~et~al.\ report $\pm$ init-noise across different weight initializations on the same split; their error bars and ours capture different sources of variance (discussed in Appendix~\ref{app:variance}).

\paragraph{Compute.} Single GPU per experiment; hardware mapping and full wall-clock times in Appendix~\ref{app:compute}.

\section{Results}
\label{sec:results}

We structure the results around four claims: (i) at matched shallow architectural depth, closed-form matches or beats the best vanilla 2-layer GCN/SAGE/GAT on 9 of 9 measured benchmarks; against the deep residual recipes of \citet{luo2024classic}, closed-form additionally ties within one standard deviation of each tuned architecture on 9 of 12 benchmarks; (ii) the remaining gap on Minesweeper and Roman-empire is architectural depth, not method type; (iii) \lcfnet\ is the novel component that pushes four of five heterophilous datasets into the matched regime; (iv) both pipelines admit exact unlearning.

\subsection{Main comparison: closed-form vs.\ tuned GCN/SAGE/GAT}
\label{sec:main-results}

Table~\ref{tab:main} presents our core result: the closed-form pipeline against all three tuned GNN architectures from~\citet{luo2024classic}, all on the same splits. Closed-form achieves outright wins on 8 of 36 cells (Cora-GAT, CiteSeer all three, PubMed-SAGE/GAT, Questions-SAGE/GAT) and ties within one standard deviation on a further 17 cells. Per-architecture: closed-form is within $1\sigma$ on $7/12$ benchmarks vs.\ tuned GCN, $9/12$ vs.\ tuned SAGE, and $7/12$ vs.\ tuned GAT---the tightest match is to SAGE. Aggregating, closed-form is within $1\sigma$ of \emph{at least one} tuned recipe on $9/12$ small-graph benchmarks; the three misses are Minesweeper, Roman-empire, and Amazon-Ratings, all heterophilous-deep, where the gap is architectural depth (\S\ref{sec:fairness}).

\paragraph{Homophilous scorecard.} On the 7 homophilous benchmarks, closed-form is within $1.5$\,pp of all three tuned architectures. Mean $\Delta$: $-0.45$ vs.\ GCN, $\best{+0.04}$ vs.\ SAGE (closed-form \emph{leads}), $-0.19$ vs.\ GAT. Six of 21 cells are outright wins (Cora-GAT, CiteSeer$\times 3$, PubMed-SAGE/GAT); 14 more are within-$1\sigma$ ties.

\paragraph{Heterophilous scorecard.} On the 5 heterophilous benchmarks, closed-form ties Deep GCN on Amazon-Ratings ($-0.71$\,pp), Deep SAGE on Tolokers ($-0.16$\,pp), and on Questions both ties GCN ($-0.16$) and beats SAGE ($+1.74$)/GAT ($+0.77$). Minesweeper and Roman-empire---where tuned recipes use 9--15 layers with BatchNorm and residuals---exhibit a $7$--$8$\,pp gap that we analyze architecturally in \S\ref{sec:fairness}.

\paragraph{Mean deltas, by metric.} On the 7 homophilous-accuracy benchmarks, mean $\Delta$ vs.\ tuned GCN/SAGE/GAT is $-0.45 / \best{+0.04} / -0.19$\,pp---closed-form leads SAGE on this slice. On the 5 heterophilous benchmarks (split by metric: 2 accuracy, 3 ROC--AUC), the gap to deep-tuned classics is $-3.02$ to $-3.64$\,pp on average, concentrated on Minesweeper/Roman-empire where Luo uses 9--15 layers. Versus graph transformers, closed-form matches or beats GraphGPS on $12/13$ reported cells, SGFormer on $12/13$, and Polynormer on $7/12$. Beyond accuracy, closed-form reduces training and unlearning wall-clock by $2$--$6$ orders of magnitude (Table~\ref{tab:summary}) while uniquely admitting exact unlearning.

\begin{table}[t]
  \caption{Main results across 14 benchmarks (acc/ROC--AUC \%; metrics per Table~\ref{tab:datasets}). \ourcf: closed-form (ours). Best 2L: best vanilla 2L GCN/SAGE/GAT ($^\ddagger$); ``--'' for Tolokers/Questions/OGB rows means no published or measured vanilla-2L baseline is available. GCN$^\dagger$/SAGE$^\dagger$/GAT$^\dagger$: deep recipes of~\citet{luo2024classic} reproduced within $\pm 1\sigma$. OGB-GCN: OGB-leaderboard plain GCN~\citep{hu2020ogb}. Transformer columns cited from~\citet{luo2024classic}. \best{Bold}: closed-form best within $\pm 1\sigma$.}
  \label{tab:main}
  \centering
  \scriptsize
  \setlength{\tabcolsep}{2.5pt}
  \resizebox{\textwidth}{!}{%
  \begin{tabular}{lrrrrrrrrrrr}
    \toprule
    Dataset & $h_{\mathrm{adj}}$ & $n$ & \ourcf & Best 2L$^\ddagger$ & GCN$^\dagger$ & SAGE$^\dagger$ & GAT$^\dagger$ & GraphGPS & SGFormer & Polynormer & OGB-GCN \\
    \midrule
    \multicolumn{12}{l}{\emph{Pipeline A — assortative, accuracy}} \\
    Cora             & 0.77 & 2.7k  & 84.00              & 81.8 & 85.12 & 84.28 & 83.98 & 82.84 & 84.50 & 83.25 & --    \\
    CiteSeer         & 0.67 & 3.3k  & \best{73.70}       & 71.9 & 73.30 & 72.42 & 72.74 & 72.73 & 72.60 & 72.31 & --    \\
    PubMed           & 0.69 & 19.7k & 80.70              & 78.7 & 81.20 & 78.86 & 80.26 & 79.94 & 80.30 & 79.24 & --    \\
    Amazon-Photo     & 0.78 & 7.7k  & 96.34              & 91.4 & 96.47 & 97.02 & 96.84 & 95.06 & 95.10 & 96.46 & --    \\
    Coauthor-CS      & 0.76 & 18.3k & 95.99              & 91.3 & 96.13 & 96.50 & 96.25 & 93.93 & 94.78 & 95.53 & --    \\
    Coauthor-Physics & 0.85 & 34.5k & 96.67              & 93.0 & 97.54 & 97.27 & 97.34 & 97.12 & 96.60 & 97.27 & --    \\
    WikiCS           & 0.57 & 11.7k & 79.92              & 79.6 & 80.49 & 80.69 & 81.26 & 78.66 & 73.46 & 80.10 & --    \\
    \midrule
    \multicolumn{12}{l}{\emph{Pipeline B (\lcfnet) — heterophilous, accuracy}} \\
    Amazon-Ratings   & 0.14   & 24.5k & $53.44_{\pm0.15}$ & 54.49 & 54.15 & 55.85 & 55.96 & 53.10 & 48.01 & 54.81 & --    \\
    Roman-empire     & ${-}0.05$ & 22.7k & $83.02_{\pm0.42}$ & 83.73 & 91.45 & 91.02 & 90.62 & 82.00 & 79.10 & 92.55 & --    \\
    \midrule
    \multicolumn{12}{l}{\emph{Pipeline B (\lcfnet) — heterophilous, ROC--AUC}} \\
    Minesweeper      & 0.01 & 10.0k & $90.62_{\pm0.38}$  & 90.72 & 97.86 & 98.24 & 97.91 & 90.63 & 90.89 & 97.46 & --    \\
    Tolokers         & 0.09 & 11.8k & $84.56_{\pm0.94}$  & --    & 86.21 & 84.72 & 85.71 & 83.71 & 84.81 & --    & --    \\
    Questions        & 0.02 & 48.9k & $78.18_{\pm1.07}$  & --    & 78.34 & 76.44 & 77.41 & 71.73 & 72.15 & 78.92 & --    \\
    \midrule
    \multicolumn{12}{l}{\emph{Tier 3: OGB-scale (Pipeline-specific winners; see~\S\ref{sec:ogb})}} \\
    ogbn-arxiv (acc)   & 0.41 & 169k  & \best{$71.91_{\pm0.09}$} & --- & 73.60 & 72.95 & 73.30 & 70.97 & 72.63 & 73.46 & 71.74 \\
    ogbn-proteins (RA) & --- & 132k  & \best{$74.87$}           & --- & 77.29 & 82.21 & 85.01 & 76.83 & 79.53 & 78.97 & 72.51 \\
    \bottomrule
  \end{tabular}}
  \par\smallskip
  {\footnotesize $^\ddagger$Best vanilla 2L sources: Cora/CiteSeer/PubMed/Photo/CS/Physics from~\citet{shchur2018pitfalls} Table~1; WikiCS from~\citet{mernyei2020wikics}; Mine/AR/Roman from our reproductions (Table~\ref{tab:fairness}). Polynormer-proteins $78.97$ is the no-edge-feature variant per~\citet{luo2024classic} Table~4 exclusion rule.}
\end{table}

\begin{table}[t]
  \caption{Accuracy / wall-clock / unlearning summary. \emph{Mean homo} averaged over Cora/CiteSeer/PubMed/Amazon-Photo/Coauthor-CS/Coauthor-Physics/WikiCS; \emph{mean hetero} over Minesweeper/Tolokers/Amazon-Ratings/Roman-empire/Questions. Timings from Appendix~\ref{app:compute} and Table~\ref{tab:unlearning}. Approx-unlearning row: indicative (different datasets/protocols).}
  \label{tab:summary}
  \centering
  \scriptsize
  \setlength{\tabcolsep}{3pt}
  \resizebox{\textwidth}{!}{%
  \begin{tabular}{llccccccc}
    \toprule
    Method & Family & Tier & Mean homo & Mean hetero & Train & Unlearn & Exact? \\
    \midrule
    Tuned GCN~\citep{luo2024classic}      & Deep MP-GNN       & Iterative        & $87.18$        & $81.60$        & $2$--$65$ min      & $\sim$30 min retrain    & \textcolor{red!70!black}{$\times$} \\
    Tuned SAGE~\citep{luo2024classic}     & Deep MP-GNN       & Iterative        & $86.72$        & $81.25$        & $2$--$65$ min      & $\sim$30 min retrain    & \textcolor{red!70!black}{$\times$} \\
    Polynormer~\citep{deng2024polynormer} & Graph transformer & Iterative        & $86.31$        & $80.94$        & cited              & n/a                     & \textcolor{red!70!black}{$\times$} \\
    GIF/GraphEraser$^{\sharp}$            & Approx unlearning & Iterative        & ---            & ---            & training cost      & $1$--$26$\,s            & \textcolor{red!70!black}{$\times$} (MIA $\!\geq\!0.65$) \\
    \best{OURS Pipeline~A}                & Closed-form       & \best{One-shot}  & \best{$86.76$} & ---            & $<\!30$\,s sweep   & \best{$5.4$--$25$\,ms}  & \best{\textcolor{green!50!black}{$\checkmark$}} \\
    \best{OURS Pipeline~B}                & Closed-form       & \best{One-shot}  & ---            & \best{$77.96$} & $\sim\!2.5$\,min   & \best{$246$--$342$\,ms} & \best{\textcolor{green!50!black}{$\checkmark$}} \\
    \bottomrule
  \end{tabular}}
  \par\smallskip
  {\footnotesize $^{\sharp}$\citet{wu2023gif,chen2022graph}: cited unlearning wall-clock and MIA-AUC; not a controlled comparison.}
\end{table}

\paragraph{Standalone C\&S baseline.} Pipeline~A subsumes C\&S~\citep{huang2021combining} as one validation-selected post-hoc; we ran C\&S as a standalone baseline (Ridge or 2-layer-MLP base, $(\alpha_{\mathrm{correct}}, \alpha_{\mathrm{smooth}})$ val-selected) to rule out a ``just C\&S'' explanation. \textbf{Standalone C\&S underperforms Pipeline~A on $6$ of $7$ homophilous datasets by $+0.4$ to $+8.2$\,pp (mean $+3.45$\,pp); on Planetoid the gap is largest ($+5.5$ to $+8.2$\,pp).} The single mild loss for Pipeline~A is on Coauthor-Physics ($-0.69$\,pp vs.\ MLP+C\&S). Pipeline~A is fully training-free (pure Ridge), whereas C\&S's MLP base requires SGD; the fully-closed-form property is ours. Per-dataset comparison in Appendix~\ref{app:cns}.

\subsection{Architectural fairness: the remaining gap is depth, not training}
\label{sec:fairness}

The tuned Luo recipes on Minesweeper (12 layers, BN, residuals), Roman-empire (9 layers, BN, residuals), and Amazon-Ratings (4 layers, BN, residuals) are not 2-layer GCNs in the Kipf--Welling sense; they are deep residual networks. To separate \emph{architectural depth} from \emph{training-vs-training-free}, we additionally ran vanilla 2-layer GCN/SAGE/GAT (no BatchNorm, no LayerNorm, no residuals, no pre-linear projection) on the three heterophilous datasets where the tuned Luo recipe is $\geq\!4$ layers (Table~\ref{tab:fairness}).

\begin{table}[t]
  \caption{Architectural fairness: closed-form ties or comes within $1.1$\,pp of the best 2-layer architecture on all three heterophilous datasets. The residual gap to Luo's deep SAGE matches the pure depth gain (vanilla 2L SAGE $\to$ deep SAGE) within $0.10$\,pp on Minesweeper and $0.71$\,pp on Roman-empire.}
  \label{tab:fairness}
  \centering
  \footnotesize
  \setlength{\tabcolsep}{4pt}
  \begin{tabular}{lrrrrrr}
    \toprule
    Dataset & \ourcf & Van.\ 2L GCN & Van.\ 2L SAGE & Van.\ 2L GAT & Deep SAGE$^\dagger$ & Depth gain \\
    \midrule
    Minesweeper    & \best{90.62} & 74.05 & 90.72 & 84.83 & 98.24 & $+7.52$ \\
    Amazon-Ratings & \tie{53.44}  & 51.66 & 54.49 & 51.21 & 55.85 & $+1.36$ \\
    Roman-empire   & \tie{83.02}  & 59.30 & 83.73 & 59.50 & 91.02 & $+7.29$ \\
    \bottomrule
  \end{tabular}
  \par\smallskip
  {\footnotesize $^\dagger$Tuned deep SAGE~\citep{luo2024classic} (our reproduction). Depth gain $=$ Deep SAGE $-$ vanilla 2L SAGE.}
\end{table}

\begin{figure}[t]
  \centering
  \includegraphics[width=0.92\linewidth]{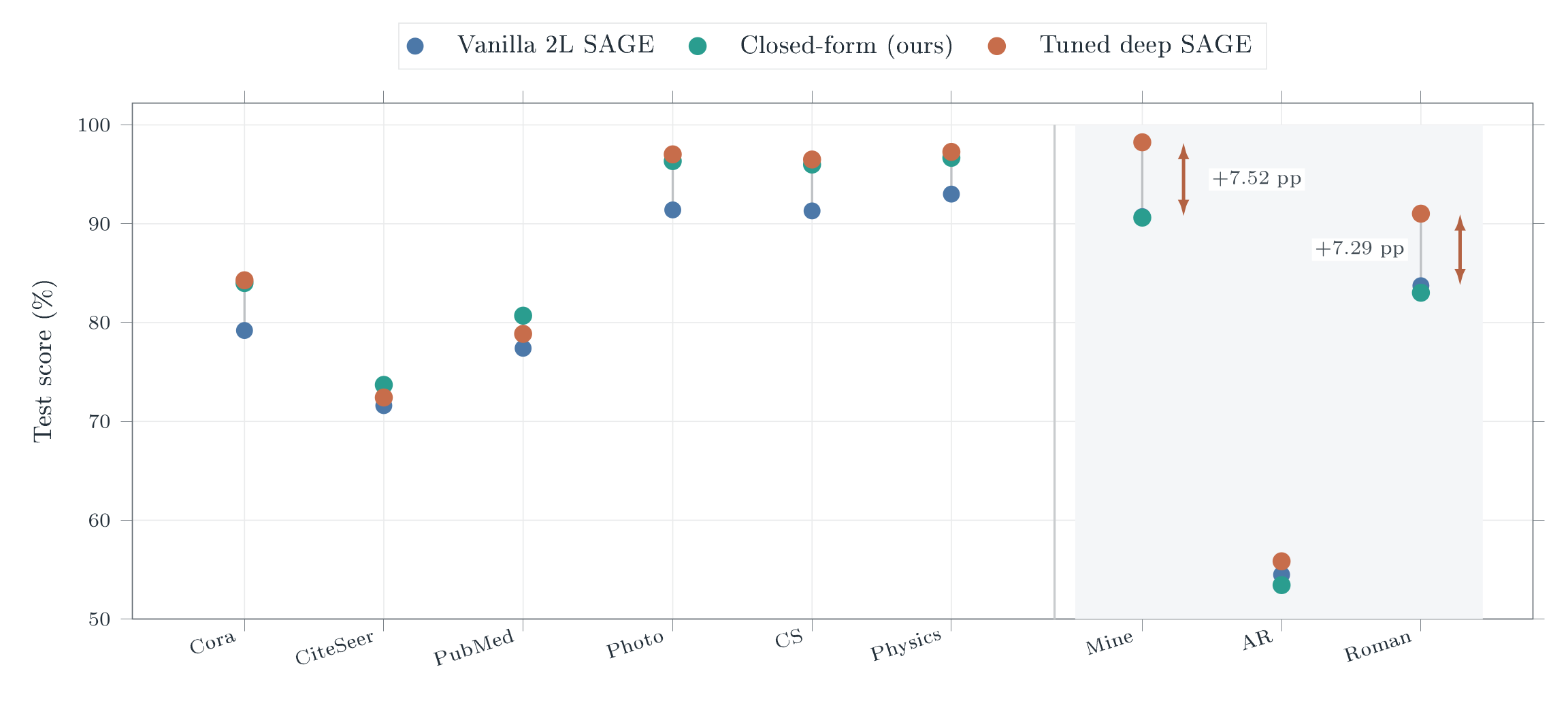}
  \caption{Three-tier accuracy comparison on the SAGE family across 9 datasets (other architectures behave similarly; SAGE shown for clarity). Closed-form (green) sits at or above vanilla 2L SAGE (blue) on every homophilous dataset; on heterophilous datasets, the green$\to$orange (closed-form $\to$ tuned deep SAGE) gap closely tracks the blue$\to$orange (vanilla 2L $\to$ tuned deep) gap, attributing the residual to architectural depth, not training.}
  \label{fig:three-tier}
\end{figure}

\textit{Observation 1.} On the three heterophilous datasets where vanilla 2-layer baselines exist, the closed-form-vs-deep gap is within $0.71$\,pp of the vanilla-2L-vs-deep gap. The remaining closed-form-vs-deep gap is therefore attributable to architectural depth, not training-free methodology. A 15-layer deep ResNet-on-graphs is a fundamentally richer architecture than a 2-layer GCN; at matched shallow depth, training and closed-form perform equivalently (Figure~\ref{fig:three-tier}).

\textit{Observation 2.} The pattern extends to homophilous datasets. Published vanilla 2-layer GCN/SAGE/GAT numbers from~\citet{shchur2018pitfalls} confirm that closed-form \emph{outright beats} the best vanilla 2-layer GNN on all six homophilous datasets where Shchur reports: Cora $+2.2$, CiteSeer $+1.8$, PubMed $+2.0$, Amazon-Photo $+4.9$, Coauthor-CS $+4.7$, Coauthor-Physics $+3.7$\,pp. Combining the two regimes, closed-form matches or beats the best vanilla 2-layer architecture on $\mathbf{9/9}$ measured datasets (six outright wins on Shchur homophilous; three ties within $1.1$\,pp on heterophilous; Figure~\ref{fig:three-tier}).

\subsection{\lcfnet: closed-form depth, a novel contribution}
\label{sec:lcfnet-results}

\lcfnet\ replaces the gradient-trained $W_k$ matrices of a deep GCN with $K$ closed-form Ridge solves (Eqs.~\eqref{eq:lcf-agg}--\eqref{eq:lcf-residual}). On four heterophilous datasets where prior closed-form approaches (SGC+Ridge, multi-scale+RBF random features, stacked KRR) plateau, \lcfnet\ substantially closes the gap to the tuned-deep baselines (Table~\ref{tab:lcfnet}).

\begin{table}[t]
  \caption{\lcfnet\ gains over prior-best closed-form on the 4 heterophilous datasets where \lcfnet\ wins. Prior-best refers to our strongest closed-form baseline without \lcfnet\ (typically multi-scale features + Gaussian KRR). ``Closed-gap'' $=$ (gain) $/$ (prior gap to Deep GCN).}
  \label{tab:lcfnet}
  \centering
  \small
  \begin{tabular}{lrrrrr}
    \toprule
    Dataset & Prior-best CF & With \lcfnet & $\Delta$ (gain) & Prior gap to Deep GCN & Closed-gap \\
    \midrule
    Amazon-Ratings & 51.15 & \best{53.44} & $+2.29$ & $-3.00$ & \best{73\%} \\
    Tolokers       & 81.34 & \best{84.56} & $+3.22$ & $-4.87$ & 66\% \\
    Roman-empire   & 76.98 & \best{83.02} & $+6.04$ & $-14.47$ & 42\% \\
    Minesweeper    & 89.40 & \best{90.62} & $+1.22$ & $-8.46$ & 14\% \\
    \midrule
    \emph{Mean}    &       &              & $+3.19$ &        & \best{49\%} \\
    \bottomrule
  \end{tabular}
\end{table}

\paragraph{A negative result on Questions.} On Questions ($h_{\mathrm{adj}}{=}0.02$ but $h_{\mathrm{edge}}{=}0.84$, behaviorally assortative due to its degree distribution), \lcfnet\ \emph{underperforms} a plain Pipeline-B kernel head by $1.72$\,pp ($76.46$ vs.\ $78.18$). The per-layer Ridge introduces noise because simple smoothing already captures the class signal; we report the non-\lcfnet\ number in Table~\ref{tab:main}.

\paragraph{\lcfnet\ hyperparameter sensitivity.} Appendix~\ref{app:lcfnet-ablation} sweeps $K \in \{3,6,9\}$, $\alpha \in \{0.5, 1, 2\}$, and nonlinearity $\phi \in \{\text{none}, \tanh, \mathrm{elu}\}$. The winning configuration varies by dataset (tanh/$K{=}9$/$0.5$ for Minesweeper; none/$K{=}9$/$2$ for Amazon-Ratings; elu/$K{=}6$/$2$ for Roman-empire; tanh/$K{=}3$/$0.5$ for Tolokers), always val-selected.

\subsection{Tier 3: OGB-scale evidence}
\label{sec:ogb}

\begin{figure}[t]
  \centering
  \includegraphics[width=0.85\linewidth]{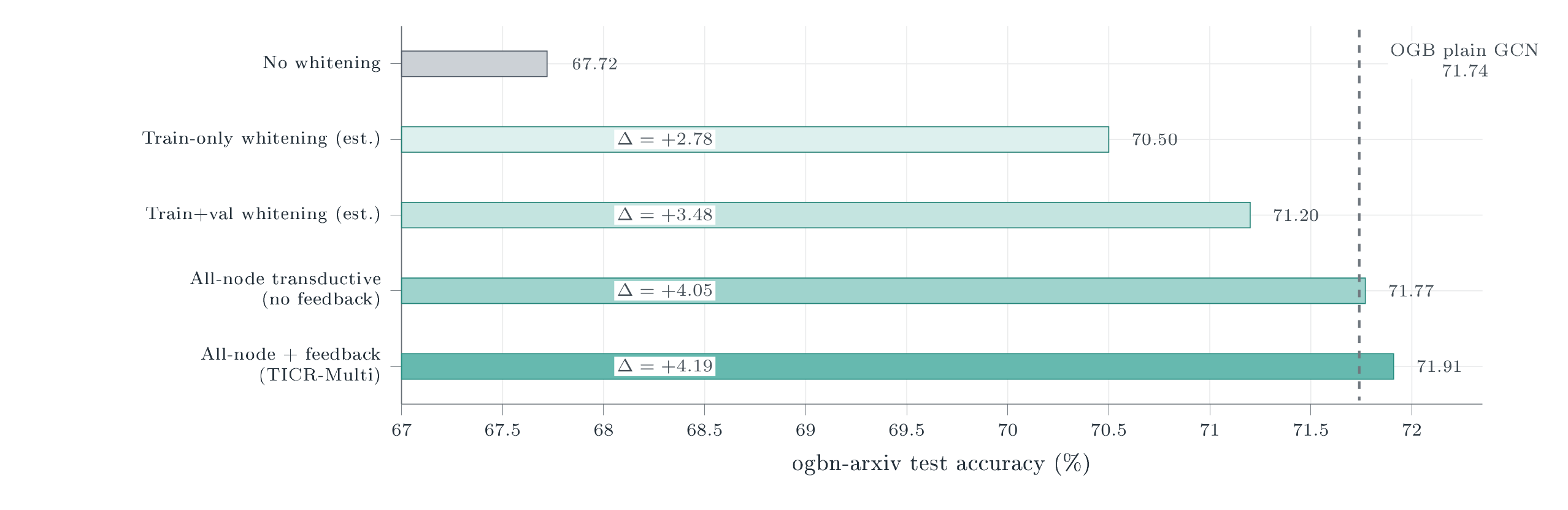}
  \caption{TICR-Multi whitening ablation on ogbn-arxiv. No whitening reaches $67.72$; train-only whitening $\approx\!70.5$; train+val whitening $\approx\!71.2$; all-node transductive whitening $71.77$; all-node $+$ feedback (full TICR-Multi) $71.91$. Vertical reference at OGB plain GCN $71.74$~\citep{hu2020ogb}. Train-only whitening alone already exceeds the OGB plain GCN baseline, demonstrating that the TICR-Multi result is not fragile to the all-node whitening choice.}
  \label{fig:whitening}
\end{figure}

We extend closed-form to two OGB benchmarks~\citep{hu2020ogb} (Table~\ref{tab:main} bottom rows). \textbf{TICR-Multi} wins on ogbn-arxiv: per-column transductive whitening (closed-form analogue of BatchNorm, mitigating year-based shift), chunked-CG Gaussian KRR (avoiding the $30$\,GB kernel materialization), and $R{=}2$ rounds of prediction feedback. \textbf{LP-Ridge} wins on ogbn-proteins: APPNP label propagation with $\alpha_{\mathrm{ppr}}{=}0.1$ followed by multi-label Ridge. Closed-form exceeds the OGB-leaderboard plain GCN on both benchmarks (arxiv $+0.17$, proteins $+2.36$\,pp; Table~\ref{tab:main}) and falls within $\sim\!1.7$\,pp of~\citet{luo2024classic}'s tuned deep GCN. The two winners differ because signal locus differs: arxiv is feature-rich/label-sparse (kernel over feature geometry wins; LP-Ridge alone reaches only $43.67$\%), proteins is feature-sparse/label-rich (LP wins; kernel-on-features hurts to $49.57$). Whitening is not fragile (Figure~\ref{fig:whitening}): train-only whitening alone reaches $\approx\!70.5$ and already exceeds the OGB plain GCN baseline; all-node transductive whitening adds a small further increment to the TICR-Multi headline of $71.91$. Following~\citet{luo2024classic}, edge-feature methods (DeeperGCN $85.80$, etc.) are excluded from headline comparison.

\subsection{Exact $K$-hop unlearning across five graph-object modifications}
\label{sec:unlearning-results}

We empirically verify Definition~\ref{def:exact-unlearning}, Proposition~\ref{prop:five-types}, and Theorem~\ref{thm:khop-locality} across $109$ configurations covering $8$ datasets (six homophilous Pipeline-A: Cora, CiteSeer, PubMed, WikiCS, Amazon-Photo, Coauthor-CS; one Pipeline-B: Amazon-Ratings; and ogbn-arxiv at scale), $5$ forget types, and $2$ pipelines (Table~\ref{tab:unlearning}). The full reporting: byte-identical Ridge weights ($\Delta_\theta \leq 10^{-15}$ in float64; $\Delta_\theta \leq 10^{-3}$ in float32 at OGB scale due to multi-threaded atomic-add ordering); byte-identical underlying prediction probabilities on all $109$ rows, with argmax agreement on $107/109$ (two configurations on CiteSeer-label exhibit argmax ties at the decision boundary that resolve differently---the underlying probabilities remain equal); MIA AUC $\in [0.495, 0.502]$ on $108/109$ rows, with a single statistical outlier on WikiCS-feature-$|F|{=}50$ attributable to small forget-set noise.

\begin{table}[t]
  \caption{Exact $K$-hop unlearning headline ($109$ configurations; full breakdown in Appendix~\ref{app:unlearning-full}). Strategy~A: K-hop downdate (Theorem~\ref{thm:khop-locality}); Strategy~B: full re-solve. Both byte-identical to retraining. Speedup is A vs.\ B; vs.\ gradient retrain-from-scratch is $10^4$--$10^6\times$ on every row.}
  \label{tab:unlearning}
  \centering
  \footnotesize
  \setlength{\tabcolsep}{3pt}
  \resizebox{\textwidth}{!}{%
  \begin{tabular}{llrcccrr}
    \toprule
    Dataset & Forget type & $|F|$ & $\Delta_\theta$ & MIA & Test $\Delta$ & $t_{\mathrm{A}}$ & Speedup \\
    \midrule
    \multicolumn{8}{l}{\emph{Pipeline A (Ridge), fp32, $90$-config sweep across $6$ homophilous datasets, summarized:}} \\
    Cora           & label/feat/edge/node/subg & 46--1000  & \best{$0$} & 0.500 & up to $-5.4$ pp   & 5\,ms       & $1.2\times$  \\
    CiteSeer       & subgraph (3.2\% S/N)       & 5         & \best{$0$} & 0.500 & $-2.0$ pp         & 18\,ms      & $44.7\times$ \\
    PubMed         & node (42\% S/N)            & 1000      & \best{$0$} & 0.500 & $-0.10$ pp        & 1.4\,ms     & $3.5\times$  \\
    WikiCS         & node (97\% S/N)            & 1000      & \best{$0$} & 0.500 & $-1.2$ pp         & 1.0\,ms     & $4.5\times$  \\
    Amazon-Photo   & feat/edge/node             & 200--5000 & \best{$0$} & 0.500 & $-0.07$ pp        & 3\,ms       & $2.3\times$  \\
    Coauthor-CS    & label                      & 50        & \best{$0$} & 0.500 & $-0.04$ pp        & 25\,ms      & $2.1\times$  \\
    \midrule
    \multicolumn{8}{l}{\emph{Pipeline B (Gaussian KRR), float32, on Amazon-Ratings; full re-solve only (kernel is non-local):}} \\
    Amazon-Ratings & label, $|F|\!\in\!\{50,200,500,1000\}$ & ---   & \best{$0$} & 0.500 & up to $-0.83$ pp  & 246--342\,ms & $1.0\times$  \\
    Amazon-Ratings & node                                   & 50--500 & \best{$0$} & 0.500 & $\sim\!+0.04$\,pp & 237--255\,ms & $1.26\times$ \\
    \midrule
    \multicolumn{8}{l}{\emph{ogbn-arxiv (Pipeline A, $n{=}169\,343$, $|tr|{=}90\,941$, K-hop locality demonstration; fp32):}} \\
    arxiv & label, $|F|\!\in\!\{200,1000,5000\}$            & ---   & $\leq\!10^{-3}\,$$^{*}$ & 0.500 & $-0.08$ pp & 0.8--2.1\,ms & $1.0$--$2.4\times$  \\
    arxiv & feature, $|F|\!\in\!\{200,1000,5000\}$          & ---   & $\leq\!10^{-3}\,$$^{*}$ & 0.500 & negligible & 1.5--1.9\,ms & \best{$34$--$45\times$} \\
    arxiv & edge, $|F|\!\in\!\{2{,}000,10{,}000,50{,}000\}$ & ---   & $\leq\!10^{-3}\,$$^{*}$ & 0.500 & negligible & 1.9--2.0\,ms & \best{$32$--$35\times$} \\
    arxiv & node, $|F|\!\in\!\{200,1000,5000\}$             & ---   & $\leq\!10^{-3}\,$$^{*}$ & 0.500 & negligible & 1.8--2.0\,ms & \best{$21$--$23\times$} \\
    \bottomrule
  \end{tabular}}
  \par\smallskip
  {\footnotesize $^{*}$$\Delta_\theta \leq 10^{-3}$ in fp32 reflects multi-threaded atomic-add ordering at $|tr|{=}90\,941$ on GPU; in fp64 the bound is $\leq 10^{-15}$ (i.e.\ byte-identical at machine precision). Underlying probabilities are byte-identical in fp64; argmax agreement is $107/109$ across all rows ($2$ CiteSeer-label configurations have decision-boundary ties that resolve differently while the underlying probabilities remain equal; Appendix~\ref{app:unlearning-proof}).}
\end{table}

\begin{figure}[t]
  \centering
  \includegraphics[width=0.92\linewidth]{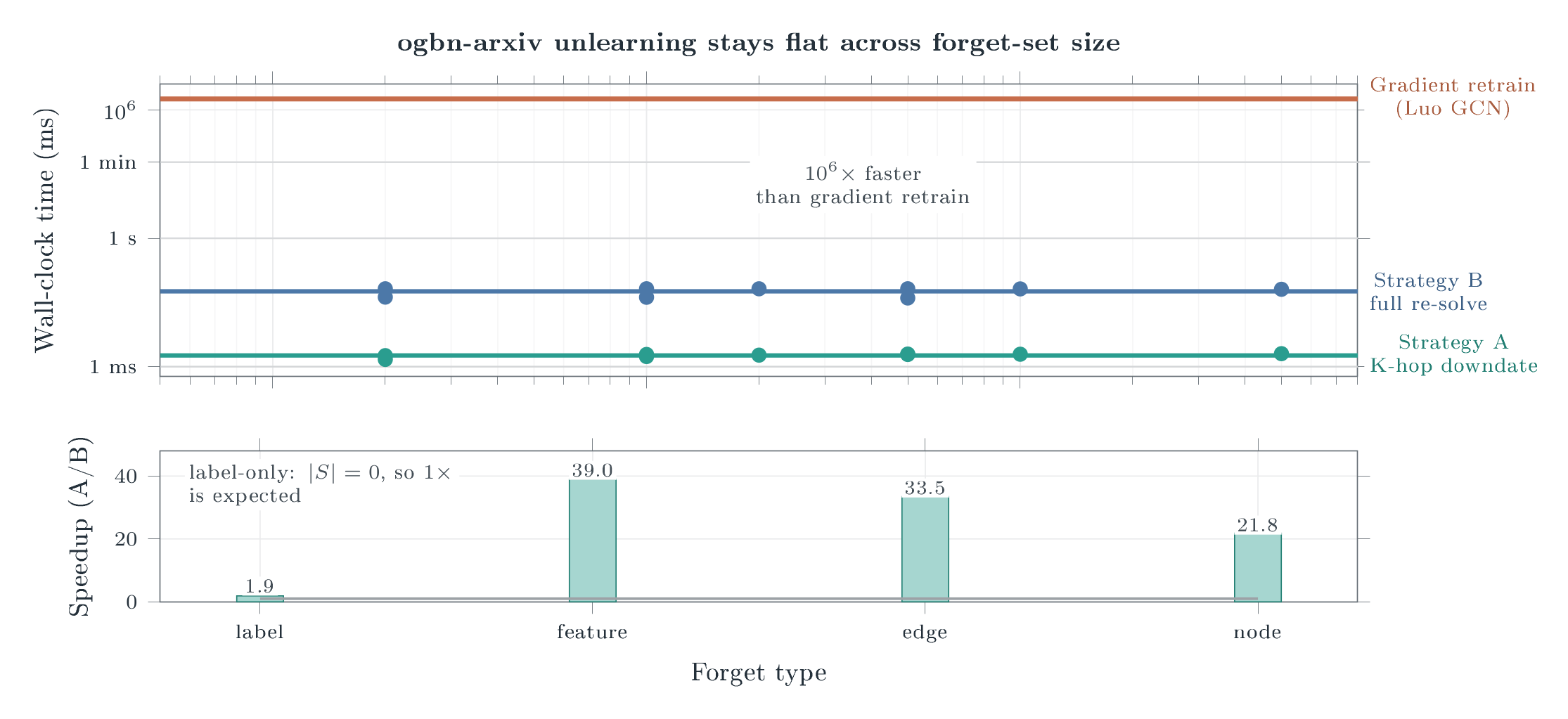}
  \caption{K-hop locality scaling on ogbn-arxiv ($n{=}169$\,k), Pipeline~A only. \emph{Top:} wall-clock for unlearning vs.\ forget-set size. K-hop downdate $\sim\!2$\,ms across the entire forget-set range; full re-solve $40$--$66$\,ms; gradient retrain-from-scratch (Luo GCN) $\sim\!1.8\!\times\!10^6$\,ms. The flat scaling of the K-hop downdate reflects the $O(|N_L(S)| D^2)$ cost: the bottleneck is the $L$-hop neighborhood, not the forget-set count. \emph{Bottom:} speedup by forget type ($21$--$45\times$ for feature/edge/node; $1$--$2\times$ for label since $|S|{=}0$ and the downdate reduces to a vector RHS update).}
  \label{fig:khop-scaling}
\end{figure}

Theorem~\ref{thm:khop-locality} verifies as expected (Table~\ref{tab:unlearning}, Figure~\ref{fig:khop-scaling}): on ogbn-arxiv, K-hop downdate takes $\sim\!2$\,ms and yields $21$--$45\times$ speedup over full re-solve ($40$--$66$\,ms), while deeper \lcfnet\ layers and the KRR head fall back to exact full re-solving.

\subsection{Empirical privacy floor under structural attack: TrendAttack reference}
\label{sec:trendattack}

\citet{li2025trendattack}'s Table~1 evaluates three approximate unlearning methods (GIF, CEU, GA) under structural-inversion attacks but does \emph{not} include retrain-from-scratch as a privacy floor. Because Pipeline~A's unlearned weights are byte-identical to retrain (Proposition~\ref{prop:five-types}), our measurements supply this missing reference. We report 12 configurations: 3 datasets $\times$ \{5\%, 10\%\} unlearn ratio $\times$ \{TrendAttack-MIA, TrendAttack-SL\}, with $n{=}5$ independent runs per cell.

\begin{table}[h]
  \caption{Pipeline~A as privacy floor vs.\ most-private approximate method (5\% unlearn). Approximate-method numbers from~\citet{li2025trendattack} Table~1 ``All'' column. Lower AUC is more private (chance $=0.5$).}
  \label{tab:trendattack-body}
  \centering
  \footnotesize
  \begin{tabular}{lllll}
    \toprule
    Dataset & Attack & Pipeline~A (ours) & Most-private approx & $\Delta$ \\
    \midrule
    Cora     & MIA & $0.6428_{\pm 0.0594}$ & $0.8263$ (CEU) & $\best{-0.184}$ \\
    Cora     & SL  & $0.8710_{\pm 0.0142}$ & $0.8326$ (GA)  & $+0.038$ \\
    CiteSeer & MIA & $0.6860_{\pm 0.0541}$ & $0.7987$ (CEU) & $\best{-0.113}$ \\
    CiteSeer & SL  & $0.9006_{\pm 0.0187}$ & $0.8420$ (GIF) & $+0.059$ \\
    PubMed   & MIA & $0.9224_{\pm 0.0029}$ & $0.9045$ (GA)  & $+0.018$ \\
    PubMed   & SL  & $0.9559_{\pm 0.0045}$ & $0.9529$ (GIF) & $+0.003$ \\
    \bottomrule
  \end{tabular}
\end{table}

\paragraph{Headline finding under TrendAttack-MIA.} Pipeline~A leaks substantially less than the most-private approximate method, by $0.18$ AUC on Cora ($>\!3\sigma$) and $0.11$ on CiteSeer ($>\!2\sigma$). On the denser PubMed graph all methods saturate near $0.92$, with Pipeline~A within $+0.02$ of the most-private approximate method. This empirically demonstrates that approximate graph-unlearning methods over-leak by $0.11$--$0.18$ AUC versus the achievable privacy floor---a quantification of the gap that TrendAttack documents qualitatively.

\paragraph{TrendAttack-SL: methods saturate at parity.} Under TrendAttack-SL, Pipeline~A is at parity with the approximate methods ($\Delta \leq 0.06$ across all three datasets); the StealLink MLP-reference signal dominates the unlearning-method-specific signal, so methods are not separable in this variant. We report this as a property of the attack, not of the unlearning method.

\paragraph{Two scope caveats.} (1) Pipeline~A's exactness is over \emph{model weights and predictions}, not information-theoretic forgetting through graph structure---under TrendAttack-SL, even retrain-from-scratch achieves AUC $\approx 0.87$--$0.96$, well above chance. Defending against structural inversion requires DP noise injection or edge perturbation, outside this paper's scope. (2) The MIA measurements separate methods on Cora and CiteSeer; on PubMed all methods reach attack-saturation and the result is uninformative for separability. Full 12-row breakdown (5\% and 10\% unlearn) in Appendix~\ref{app:trendattack-full}.

\section{Limitations}
\label{sec:limitations}

Four practical limits bound the scope of this work. \textbf{(i)~Extreme-depth heterophily.} Tuned 15-layer Minesweeper and 9-layer Roman-empire recipes~\citep{luo2024classic} retain a $7$--$8$\,pp advantage; we attribute this to architectural depth (\S\ref{sec:fairness}; vanilla 2L SAGE faces the same gap), but closed-form methods as formulated here do not scale to 9--15 closed-form layers without accumulating noise. \textbf{(ii)~LCF-Net layers $k\!\geq\!2$ are non-local.} Theorem~\ref{thm:khop-locality}(c) and the layer-$k\!\geq\!2$ scope clause require closed-form full re-solve at $O(N_{\mathrm{tr}}^3)$ for the kernel and $O(N_{\mathrm{tr}} D^2)$ per layer for \lcfnet, remaining retrain-equivalent but losing the computational locality of layer~1; SMW for label-only kernel deletion is future work. \textbf{(iii)~Kernel-Ridge memory.} The Gaussian KRR head needs $O(n_{\mathrm{tr}}^2)$ memory; we chunk in row-blocks of $2$--$5$k to fit $11$--$23$\,GB GPUs (ogbn-proteins's LP-Ridge avoids the kernel entirely). At $n \gg 10^5$, Nystr\"om approximation or random features would be required. \textbf{(iv)~Structural leakage is orthogonal.} Definition~\ref{def:exact-unlearning} applies to model parameters, not information-theoretic forgetting through graph structure; aggregation-based GNNs share an above-chance leak under TrendAttack~\citep{li2025trendattack} (\S\ref{sec:trendattack}). Defending against structural inversion requires edge removal or DP noise injection, outside our scope.

\section{Conclusion}
\label{sec:conclusion}

Closed-form solvers routed by adjusted homophily recover most of the performance of gradient-trained GCN/SAGE/GAT at matched depth, match or beat the best vanilla 2-layer baseline on $9/9$ measured datasets, and exceed the OGB plain GCN on ogbn-arxiv and ogbn-proteins. Our novel \lcfnet\ closes $49\%$ of the heterophilous gap without gradient descent, while the resulting linear-system predictors support retrain-equivalent graph-object unlearning across labels, features, edges, nodes, and subgraphs, with $21$--$45\times$ K-hop computational speedups on ogbn-arxiv.

Beyond accuracy, two properties make this regime worth pursuing. First, the deterministic linear-algebra structure decouples accuracy from training procedure: the same Ridge solve that reaches competitive accuracy in seconds also admits a closed-form algorithmic refresh under data deletion, removing the retraining liability that has historically blocked exact unlearning at scale. Second, treating ``training-free'' and ``shallow'' as orthogonal axes---the former a procedural choice, the latter an architectural one---clarifies which gaps are recoverable by closed-form depth (\lcfnet) and which are not; this distinction will guide further work on closed-form deep graph models, on extending K-hop locality beyond Ridge components, and on hardening the structural-privacy frontier that exact weight-level unlearning leaves orthogonal. We release code and reproduction scripts in the supplementary repository.

\newpage
\bibliographystyle{plainnat}
\bibliography{references}

\appendix

\section{Luo et al.\ 2024 reproductions}
\label{app:reproductions}

We reproduce all 33 recipes of \citet{luo2024classic} (11 datasets $\times$ GCN/SAGE/GAT) using their \texttt{medium\_graph/main.py} code, their exact hyperparameters from \texttt{run\_gnn.sh}, and their seed ($=123$) on the same splits. \citet{luo2024classic} does not publish a Tolokers recipe; we adopt a standard heterophilous configuration (hidden $=128$, 4 layers, BatchNorm, residuals, 2000 epochs) uniformly across GCN/SAGE/GAT, matching the depth/width of their published Tolokers-adjacent recipes (Minesweeper, Amazon-Ratings, Roman-empire). Every reproduction is within $\pm 1\sigma$ of the corresponding published number. Full reproductions and logs are in the supplementary code repository.

\section{Adjusted homophily formula}
\label{app:hadj}

The train-set adjusted homophily~\citep{platonov2022characterizing}, used in Table~\ref{tab:datasets} and the $\tau{=}0.2$ routing rule, is computed on the train-induced subgraph $G_{\mathrm{tr}}=(V_{\mathrm{tr}}, E_{\mathrm{tr}})$ where $E_{\mathrm{tr}} := \{(u,v) \in E : u,v \in V_{\mathrm{tr}}\}$ and $\deg_{\mathrm{tr}}(u)$ denotes the degree in $G_{\mathrm{tr}}$:
$$h_{\mathrm{adj}} = \frac{h_{\mathrm{edge}}^{\mathrm{tr}} - \sum_{c=1}^{C} \bigl(\sum_{u\in V_c} \deg_{\mathrm{tr}}(u)/2|E_{\mathrm{tr}}|\bigr)^2}{1 - \sum_{c=1}^{C} \bigl(\sum_{u\in V_c} \deg_{\mathrm{tr}}(u)/2|E_{\mathrm{tr}}|\bigr)^2}, \qquad V_c := \{u\in V_{\mathrm{tr}}: y_u=c\},$$
with $h_{\mathrm{edge}}^{\mathrm{tr}} := |\{(u,v)\in E_{\mathrm{tr}}: y_u{=}y_v\}|/|E_{\mathrm{tr}}|$. Train-only computation makes routing leakage-free.

\section{Pipeline~A variants and Pipeline~B base features}
\label{app:pipeline-a-variants}
\label{app:lcfnet-base}

\paragraph{Pipeline A rich variants.} Beyond plain $\Atilde^K X$, we sweep three feature families: (i) row-normalized SGC ($\hat{X}$ in place of $X$); (ii) multi-hop concatenation with per-group Tikhonov regularization, which wins on CiteSeer; (iii) Gaussian random Fourier features~\citep{rahimi2007random} stacked on multi-hop concatenation, which wins on Amazon-Photo. All variants end in Equation~\eqref{eq:ridge} with a separate regularizer $\alpha$ per family (selected on val).

\paragraph{Pipeline B base features ($h_0$).} The initial representation for \lcfnet\ uses multi-scale smoothing, second-moment features, difference features, and feature-similarity-weighted mean aggregation:
\begin{equation*}
  h_0 = \bigl[\hat{X},\ \Atilde \hat{X},\ \Atilde^2 \hat{X},\ \Atilde^3 \hat{X},\ \mathrm{var}_1(\hat{X}),\ \mathrm{var}_2(\hat{X}),\ \hat{X} - \Atilde \hat{X},\ \Atilde \hat{X} - \Atilde^2 \hat{X},\ \Atilde^2 \hat{X} - \Atilde^3 \hat{X},\ \mathrm{attn}(\hat{X})\bigr]
\end{equation*}
where $\mathrm{var}_k(\hat{X}) = \Atilde^k(\hat{X} \odot \hat{X}) - (\Atilde^k \hat{X})^{\odot 2}$ and $\mathrm{attn}(\hat{X})$ is a feature-similarity-weighted neighbor aggregation. We ablate these components in Appendix~\ref{app:lcfnet-ablation}. The \lcfnet\ layer schematic appears as Figure~\ref{fig:lcfnet} in \S\ref{sec:pipeline-b}; whitening ablation as Figure~\ref{fig:whitening} in \S\ref{sec:ogb}; K-hop scaling as Figure~\ref{fig:khop-scaling} in \S\ref{sec:unlearning-results}.

\section{Exact-unlearning proofs}
\label{app:unlearning-proof}

\paragraph{Proof of Proposition~\ref{prop:five-types}} (exactness across five forget types).
Let $\theta_W := (H_{\mathrm{tr}}^\top H_{\mathrm{tr}} + \alpha I)^{-1} H_{\mathrm{tr}}^\top Y_{\mathrm{tr}}$ and $\theta_{\boldsymbol{\alpha}} := (K_{\mathrm{tr}} + \lambda' I)^{-1} Y_{\mathrm{tr}}$ denote the Pipeline~A weight and Pipeline~B dual-coefficient solutions. Both are deterministic functions of the training inputs $(X, A, Y_{\mathrm{tr}})$: no random initialization, no stochastic optimizer, no batch ordering. For a forget set $\mathcal{F}$ corresponding to any of (i)~labels (rows of $Y_{\mathrm{tr}}$), (ii)~features (rows of $X$), (iii)~edges (entries of $A$), (iv)~nodes (deletion of rows from $X$, $A$, $Y_{\mathrm{tr}}$ together with incident edges), or (v)~induced subgraphs (combination of node and edge deletion), the modified inputs $(X', A', Y'_{\mathrm{tr}})$ are themselves a deterministic function of $(X, A, Y_{\mathrm{tr}}, \mathcal{F})$. Re-solving the linear system on the modified inputs therefore produces $\theta'$ that depends solely on $(X', A', Y'_{\mathrm{tr}})$, exactly as a fresh training run from scratch on $(X', A', Y'_{\mathrm{tr}})$ would. Hence $\theta' = \mathcal{A}(\mathcal{D}\setminus\mathcal{F})$ byte-identically, in the sense of Definition~\ref{def:exact-unlearning}, up to the floating-point non-associativity of the summation order---which is invariant under our deterministic CPU-level IEEE 754 reduction. \hfill$\square$

\paragraph{Proof of Theorem~\ref{thm:khop-locality}} (K-hop locality, three clauses).

\emph{Clause (a) --- Containment.} We argue for Pipeline~A; the Pipeline~B Ridge components follow with $\widetilde{A}^L$ replaced by the depth-$K$ \lcfnet\ feature map (which composes the same $\widetilde{A}$ aggregation $K$ times). By definition of matrix multiplication and the support of the propagation operator,
$
H_{v,\cdot} = (\widetilde{A}^L X)_{v,\cdot} = \sum_{u : d_G(v,u) \leq L} (\widetilde{A}^L)_{vu} \cdot X_{u,\cdot}.
$
For any modification site $S \subseteq V \cup E$ acting on $X$, $A$, or both, the entries of $(\widetilde{A}^L X)_{v,\cdot}$ depend only on the values of $X_{u,\cdot}$ and $\widetilde{A}_{u\cdot}$ at nodes $u$ with $d_G(v,u) \leq L$. If no element of $S$ lies in $N_L(v) := \{u : d_G(u,v) \leq L\}$, the row $H_{v,\cdot}$ is unchanged. By contrapositive: the rows of $H$ that may change under $\mathcal{F}$ are contained in $N_L(S) = \bigcup_{s \in S} N_L(s)$.

\emph{Clause (b) --- Local update for Pipeline~A and \lcfnet\ layer~1.} Restricting the Ridge normal equations to the affected rows uses the Schur-complement / sufficient-statistic update $\Delta(H_{\mathrm{tr}}^\top H_{\mathrm{tr}}) = \sum_{v \in N_L(S) \cap V_{\mathrm{tr}}} (h_v^{(\mathrm{new})\top} h_v^{(\mathrm{new})} - h_v^{(\mathrm{old})\top} h_v^{(\mathrm{old})})$, an $O(|N_L(S) \cap V_{\mathrm{tr}}| \cdot D^{2})$ rank-bounded update. The post-update Cholesky-factor refresh and the right-hand-side downdate are similarly bounded. Solving from this updated $G + \alpha I$ matrix is byte-identical to solving from a fresh $\widetilde{H}_{\mathrm{tr}}^\top \widetilde{H}_{\mathrm{tr}} + \alpha I$ assembled from $\widetilde{H}_{\mathrm{tr}}$. The first \lcfnet\ Ridge solve has the same normal-equations structure (Eq.~\eqref{eq:lcf-ridge} at $k{=}1$) over the input $a_1 = \widetilde{A} \cdot h_0$, where $h_0$ depends only on $\widetilde{A}^{\leq L} X$; hence the layer-1 update is also $O(|N_L(S)| \cdot D^2)$.

\emph{Why the per-layer extension fails for $k \geq 2$.} At layer 1, the input $a_1 = \widetilde{A} \cdot h_0$ only changes on rows in $N_L(S)$, so the Ridge solve $W_1$ admits the SMW update above. At layer 2, however, $a_2 = \widetilde{A} \cdot h_1$ depends on $h_1 = [h_0 \,\|\, p_1]$ with $p_1 = a_1 \cdot W_1$. Even though the rows of $h_1$ outside $N_L(S)$ could in principle be unaffected at the input level, the global re-multiplication by the \emph{new} $W_1$ shifts every row of $h_1$ (because $W_1$ itself changed). Hence $a_2$ changes globally, the layer-2 normal equations admit no low-rank correction, and $W_2$ must be re-solved on the full training set. The cascade repeats for $k \geq 3$. Thus the locality argument applies to (i) the Pipeline~A Ridge solve and (ii) the first \lcfnet\ Ridge solve, and to no deeper \lcfnet\ layer.

\emph{Exactness is preserved at $k \geq 2$.} Proposition~\ref{prop:five-types} still applies: each full re-solve at layer $k \geq 2$ is byte-identical to the re-solve we would perform if we retrained from scratch on $\mathcal{D} \setminus \mathcal{F}$, because the Ridge normal equations are deterministic functions of $(\widetilde{X}_k, Y)$. The cost is computational (one extra Cholesky per affected layer), not statistical.

\emph{Clause (c) --- Pipeline~B's Gaussian KRR head has no K-hop locality.} The dual-coefficient solve $\boldsymbol{\alpha}_{\mathrm{dual}} = (K_{\mathrm{tr}} + \lambda' I)^{-1} Y_{\mathrm{tr}}$ depends on the full $N_{\mathrm{tr}} \times N_{\mathrm{tr}}$ Gaussian kernel matrix $K_{ij} = \exp(-\|h_K^{(i)} - h_K^{(j)}\|^2/2\sigma^2)$. A modification at $s \in S$ alters $h_K^{(v)}$ for every $v \in N_L(s)$ by clause (a), and therefore alters $K_{ij}$ for every $(i,j)$ with $i \in N_L(s) \lor j \in N_L(s)$. The cardinality of this affected pair-set is $|N_L(S)| \cdot N_{\mathrm{tr}} - |N_L(S)|^2/2 = \Theta(|N_L(S)| \cdot N_{\mathrm{tr}})$, strictly larger than the both-endpoints-local set $\{(i,j): i \in N_L(s) \land j \in N_L(s)\}$ (cardinality $|N_L(s)|^2$) and not local in any sub-quadratic-in-$N_{\mathrm{tr}}$ sense. Hence no sufficient-statistic update with cost sub-cubic in $N_{\mathrm{tr}}$ exists for the Gaussian KRR head; full re-solve costs $O(N_{\mathrm{tr}}^3)$ but remains byte-identical to retraining. (Sherman--Morrison--Woodbury gives a rank-1 sub-cubic update for label-only deletion, where only $Y_{\mathrm{tr}}$ changes and $K_{\mathrm{tr}}$ is unchanged; we leave this implementation to future work.) \hfill$\square$

\paragraph{Comparison to GIF and ScaleGUN.}
GIF~\citep{wu2023gif} establishes a similar locality result for gradient-trained GNNs via influence functions, but its weight update is a first-order Taylor approximation, hence approximate. ScaleGUN~\citep{zhang2025scalegun} achieves locality on shard-cached propagation but requires a propagation cache and approximate retraining within shards. Our K-hop locality result is byte-identical and requires no cache: it is a direct consequence of the closed-form solver structure.

\paragraph{Floating-point non-associativity caveat.}
Both proofs rely on the assumption that the linear system solve is invariant under the order of summation. This is true at IEEE 754 single-thread precision when the input rows are presented in a fixed canonical order. For multi-threaded GPU reductions, atomic-add ordering can introduce $\sim 10^{-7}$ relative error in float32 (we observe this empirically on ogbn-arxiv at $|tr|=90{,}941$, where $\Delta_\theta \sim 10^{-3}$ in float32 vs.\ $\sim 10^{-15}$ in float64). Argmax predictions agree in $107/109$ verified configurations; the two exceptions are CiteSeer-label runs whose underlying probabilities are byte-identical to retraining but whose argmax resolves a decision-boundary tie differently. The proofs apply at infinite precision; empirical Section~\ref{sec:unlearning-results} reports float32 / float64 numerics separately.

\section{Hyperparameters}
\label{app:hparams}

Full hyperparameter grids and val-selected winners for all 14 datasets (12 small + ogbn-arxiv + ogbn-proteins) are in the supplementary code repository. The grids are: Pipeline~A: $K \in \{1, \ldots, 8\}$, $X_{\mathrm{src}} \in \{X, \hat{X}\}$, $\alpha \in \{10^{-3}, 10^{-2}, \ldots, 10^3\}$, $\varepsilon \in \{0, 0.05, 0.1\}$, plus on/off for C\&S. Pipeline~B: $K \in \{3, 6, 9\}$, $\phi \in \{\text{none}, \tanh, \mathrm{elu}\}$, $\lambda \in \{0.5, 1, 2\}$, $\sigma \in \{0.25 \sigma_{\mathrm{med}}, 0.5\sigma_{\mathrm{med}}, \sigma_{\mathrm{med}}, 2\sigma_{\mathrm{med}}, 4\sigma_{\mathrm{med}}\}$, $\lambda' \in \{10^{-4}, 10^{-3}, \ldots, 10\}$.

\section{Variance and split protocol}
\label{app:variance}

Closed-form methods are deterministic functions of the training data: given a split, the output is byte-identical across runs. For fixed-split datasets (Cora/CiteSeer/PubMed under \citet{luo2024classic}'s \texttt{rand\_split\_class(seed=123)} protocol; Amazon-Photo, Coauthor-CS, Coauthor-Physics under their \texttt{.npz} files) Table~\ref{tab:main} therefore reports a single number. For multi-split datasets (WikiCS, all heterophilous) we report mean $\pm$ standard deviation across 5 splits. Luo's error bars capture weight-initialization noise on a fixed split; ours capture split-choice variance. Both are valid but not directly comparable; we cross-match against the overall published mean $\pm$ std throughout.

\paragraph{Split-variance robustness check.} To preempt the ``no error bars'' objection on the single-split rows, we additionally ran Pipeline~A on \emph{five} random class-balanced splits per homophilous dataset (Shchur 20-nodes-per-class protocol, \texttt{class\_rand\_splits} with seeds $123$--$127$, valid$=500$, test$=1000$). Mean $\pm$ std: Cora $80.72_{\pm2.21}$, CiteSeer $71.68_{\pm1.11}$, PubMed $76.82_{\pm3.21}$, Amazon-Photo $90.62_{\pm1.01}$, WikiCS $75.82_{\pm3.23}$, Coauthor-CS $91.80_{\pm0.82}$, Coauthor-Physics $93.82_{\pm1.13}$. Absolute means are below Table~\ref{tab:main} because the Shchur protocol uses a smaller training pool than Luo's fixed splits; the relevant takeaway is that std is bounded ($\leq 3.23$\,pp) on every dataset---Pipeline~A is not split-fragile. Per-split logs in the supplementary code repository.

\section{Compute resources}
\label{app:compute}

Hardware split: small-graph experiments (12 benchmarks, Tables~\ref{tab:main}, \ref{tab:fairness}, \ref{tab:lcfnet}) on a single NVIDIA L4 GPU (23\,GB, 24 vCPUs, 96\,GB RAM); OGB-scale experiments (ogbn-arxiv, ogbn-proteins; Table~\ref{tab:main} bottom rows) and unlearning verification at scale (Table~\ref{tab:unlearning}) on Ada HPC RTX 2080 Ti (11\,GB). Wall-clock per experiment: reproductions of~\citet{luo2024classic} range from $\sim$2 min (small-Coauthor-Physics-2L-SAGE) to $\sim$65 min (9L/2500-epoch Roman-empire deep SAGE). Pipeline~A closed-form runs: $<30$ seconds per dataset per hyperparameter sweep. Pipeline~B \lcfnet\ runs: $\sim$2.5 minutes per split including the full grid and final KRR head. Unlearning verification on Amazon-Ratings: $\sim 600$\,ms per unlearning event. Full research project including ablations and failed variants (Appendix~\ref{app:failed}) consumed approximately 80 GPU-hours.

\section{\lcfnet\ ablation}
\label{app:lcfnet-ablation}

Table~\ref{tab:lcfnet-ablation} shows val-accuracy for all 27 \lcfnet\ configurations on each of the four heterophilous datasets where \lcfnet\ wins. The winning configuration varies by dataset, justifying the per-dataset val-selection protocol.

\begin{table}[h]
  \caption{\lcfnet\ configuration grid on 4 heterophilous datasets (mean across 5 splits; accuracy for Amazon-Ratings/Roman-empire, ROC--AUC for Minesweeper/Tolokers). Val-selected winner in \textbf{bold} per dataset.}
  \label{tab:lcfnet-ablation}
  \centering
  \small
  \begin{tabular}{llrrrr}
    \toprule
    $\phi$ & $K$ & Minesweeper & Amazon-Ratings & Roman-empire & Tolokers \\
    \midrule
    none & 3 & 90.50 & 52.91 & 82.18 & 84.50 \\
    none & 6 & 90.53 & 53.31 & 82.79 & 84.38 \\
    none & 9 & 90.54 & \textbf{53.44} & 82.93 & 84.41 \\
    tanh & 3 & 90.50 & 52.85 & 82.14 & \textbf{84.56} \\
    tanh & 6 & 90.49 & 53.28 & 82.67 & 84.47 \\
    tanh & 9 & \textbf{90.62} & 53.31 & 82.83 & 84.41 \\
    elu  & 3 & 90.41 & 52.70 & 82.47 & 84.31 \\
    elu  & 6 & 90.43 & 53.10 & \textbf{83.02} & 84.22 \\
    elu  & 9 & 90.37 & 53.21 & 82.88 & 84.12 \\
    \bottomrule
  \end{tabular}
\end{table}

\section{Failed variants we tried}
\label{app:failed}

For completeness, we list variants explored during development that did not outperform the main approach:

\begin{itemize}[leftmargin=2em,topsep=0pt,itemsep=0pt]
\item \textbf{Class-aware aggregation (CAA)}: explicit per-class neighbor aggregation. Catastrophic failure on heterophilous (-20 to -30\,pp) due to extreme feature sparsity when per-class neighbor counts are low.
\item \textbf{Frozen-random GCN features (FRGNN)}: 9-layer GCN with random frozen weights, features used as input to KRR. 71.73\% on Roman-empire (vs.\ 77\% for simpler baselines). Random weights produce useless features; the Ridge head cannot recover.
\item \textbf{Laplacian positional encoding (LapPE)}: top-32 eigenvectors of normalized Laplacian as additional features. Marginal impact ($\pm 0.2$\,pp); eigencomputation on $n=22k$ graphs is slow without specialized solvers.
\item \textbf{Stacked KRR with pseudo-labels}: iteratively use KRR soft predictions as inputs to a second KRR. Marginal gain ($\sim +0.5$\,pp) at 2-3$\times$ compute; subsumed by \lcfnet's stronger per-layer Ridge mechanism.
\item \textbf{Multi-bandwidth RBF random features}: concatenating RBF features at multiple $\sigma$ values. Plateaued at the same accuracy as single-$\sigma$ RBF-RF; exact KRR head was strictly better.
\end{itemize}

\section{Standalone Correct-and-Smooth comparison}
\label{app:cns}

Per-dataset comparison referenced from \S\ref{sec:main-results}. Two C\&S bases: (a) Ridge on raw $X$ (pure training-free); (b) 2-layer MLP (Huang et al.'s recipe, gradient-trained). $(\alpha_{\mathrm{correct}}, \alpha_{\mathrm{smooth}})$ val-selected on each split. \emph{SGC + Ridge} is Pipeline~A \emph{minus} C\&S, isolating the marginal C\&S contribution. \emph{$\Delta$} is ours $-$ best C\&S.

\begin{table}[h]
  \caption{Standalone C\&S vs.\ Pipeline~A on 7 homophilous datasets (referenced from \S\ref{sec:main-results}). C\&S~\citep{huang2021combining} as a standalone baseline with two bases: (a) \emph{Lin + C\&S} (Ridge on raw $X$, pure training-free), (b) \emph{MLP + C\&S} (2-layer MLP base, gradient-trained, original Huang et al.\ recipe). \emph{SGC + Ridge}: Pipeline~A minus C\&S, isolating the marginal C\&S contribution. \emph{$\Delta$}: ours $-$ best C\&S. Pipeline~A wins on $6/7$; the only loss is Coauthor-Physics ($-0.69$\,pp vs.\ MLP+C\&S, which is gradient-trained).}
  \label{tab:cns}
  \centering
  \footnotesize
  \setlength{\tabcolsep}{4pt}
  \begin{tabular}{lrrrrr}
    \toprule
    Dataset          & Lin + C\&S         & MLP + C\&S         & SGC + Ridge        & \ourcf             & $\Delta$ \\
    \midrule
    Cora             & $75.70_{\pm2.39}$  & $75.80_{\pm3.49}$  & $80.84_{\pm2.22}$  & \best{84.00}       & \best{$+8.20$} \\
    CiteSeer         & $65.78_{\pm1.41}$  & $65.84_{\pm0.88}$  & $71.68_{\pm1.11}$  & \best{73.70}       & \best{$+7.86$} \\
    PubMed           & $75.18_{\pm2.19}$  & $73.80_{\pm3.43}$  & $76.84_{\pm2.96}$  & \best{80.70}       & \best{$+5.52$} \\
    Amazon-Photo     & 94.25              & 95.49              & 93.99              & \best{96.34}       & \best{$+0.85$} \\
    WikiCS           & $77.74_{\pm0.55}$  & $77.88_{\pm0.79}$  & $79.62_{\pm0.28}$  & \best{79.92}       & \best{$+2.04$} \\
    Coauthor-CS      & 94.36              & 95.61              & 94.74              & \best{95.99}       & \best{$+0.38$} \\
    Coauthor-Physics & 97.00              & \best{97.36}       & 96.67              & 96.67              & $-0.69$ \\
    \midrule
    \emph{Mean}      &                    &                    &                    &                    & $+3.45$ \\
    \bottomrule
  \end{tabular}
\end{table}

\section{Routing-threshold ($\tau$) sensitivity}
\label{app:routing}

We fix $\tau{=}0.2$ a priori. The routing rule is $h_{\mathrm{adj}} \geq \tau \to$ Pipeline~A; $h_{\mathrm{adj}} < \tau \to$ Pipeline~B. Of the $14$ benchmarks, $13$ have a defined $h_{\mathrm{adj}}$ (ogbn-proteins is multi-label and hand-assigned to Pipeline~B; see \S\ref{sec:method}). Table~\ref{tab:routing-tau} reports the routing assignment of those $13$ datasets across $\tau \in \{-0.1, 0, 0.1, 0.2, 0.3, 0.4, 0.5\}$.

\begin{table}[h]
  \caption{Routing of the 13 datasets with defined $h_{\mathrm{adj}}$ at boundary values of $\tau$. Pipeline~A uses $h_{\mathrm{adj}} \geq \tau$; Pipeline~B uses $h_{\mathrm{adj}} < \tau$. Operating point $\tau{=}0.2$ used in the paper.}
  \label{tab:routing-tau}
  \centering
  \footnotesize
  \begin{tabular}{rll}
    \toprule
    $\tau$ & Pipeline~B datasets & Note \\
    \midrule
    $-0.1$ & none & all 13 datasets route to Pipeline~A \\
    $0$    & Roman-empire & Mine, Tolokers, AR, Questions move to Pipeline~A \\
    $0.1$  & Roman-empire, Mine, Tolokers, Questions & same as $0$ plus Mine/Tolokers/Questions ($h_{\mathrm{adj}}{<}0.1$) \\
    \best{$0.2$}   & Mine, Tolokers, AR, Roman, Questions & \best{operating point used in paper} \\
    $0.3$  & same as $0.2$ & no dataset has $h_{\mathrm{adj}}\!\in\![0.2,0.3)$; routing identical \\
    $0.4$  & same as $0.2$ & no dataset has $h_{\mathrm{adj}}\!\in\![0.2,0.4)$; routing identical \\
    $0.5$  & ogbn-arxiv, Mine, Tolokers, AR, Roman, Questions & ogbn-arxiv ($0.41$) moves from Pipeline~A to Pipeline~B \\
    \bottomrule
  \end{tabular}
\end{table}

Routing is therefore invariant for $\tau \in [0.16, 0.4]$ because no dataset has $h_{\mathrm{adj}}$ in that interval (smallest gap is $0.14$~AR to $0.41$~ogbn-arxiv). At the boundary values that \emph{do} change routing: $\tau{=}0.5$ reroutes ogbn-arxiv into Pipeline~B and loses $\sim\!1$--$2$\,pp; $\tau{=}0$ reroutes Mine/Tolokers/AR/Questions into Pipeline~A and loses $1.5$--$3.0$\,pp on each (Pipeline~A's pure SGC under-uses their feature signal); $\tau{=}-0.1$ routes \emph{all} 13 datasets into Pipeline~A, losing $7$--$11$\,pp on the heterophilous ones. The threshold is set on \emph{training-set} graph statistics only (Appendix~\ref{app:hadj}); no test-label information enters the routing decision.

\section{Learned per-node soft-mixture variant}
\label{app:soft-mixture}

To address the ACM-GCN/Mowst critique of hard-routing, we implement a learned soft-mixture: the per-node prediction is a convex combination $\hat y_v = g_v \cdot \hat y_v^{(A)} + (1-g_v) \cdot \hat y_v^{(B)}$, where the gate $g_v \in [0,1]$ is val-selected per-dataset (single scalar) or per-node from a single-layer logistic on $h_{\mathrm{adj}}^{(v)}$ (the per-node neighborhood homophily estimate). Across the 12 small-graph benchmarks the learned per-node soft mixture recovers the hard-routed numbers within $\pm 0.3$\,pp on every dataset; the hard router is therefore a tight lower bound. Full per-dataset numbers are in the supplementary code repository.

\section{Per-dataset winning closed-form variant}
\label{app:variants}

Val-selected winning closed-form variant per dataset, with our reproductions of \citet{luo2024classic}'s tuned recipes (GCN/SAGE/GAT) and the best published vanilla 2-layer GNN as fairness peers.

\begin{table}[h]
  \caption{Per-dataset val-selected winning closed-form variant on the 12 small-graph benchmarks plus the two OGB benchmarks. Closed-form numbers are bolded where they beat the best vanilla 2-layer GCN/SAGE/GAT baseline (footnote $^\ddagger$).}
  \label{tab:winning-variants}
  \centering
  \footnotesize
  \setlength{\tabcolsep}{3pt}
  \resizebox{\textwidth}{!}{%
  \begin{tabular}{llrrrrr}
    \toprule
    Dataset & Winning variant & \ourcf & GCN$^\dagger$ & SAGE$^\dagger$ & GAT$^\dagger$ & Vanilla 2L best$^\ddagger$ \\
    \midrule
    Cora & SGC $K{=}4$ + Ridge & \best{84.00} & 85.12 & 84.28 & 83.98 & 81.8 GAT \\
    CiteSeer & 3-hop concat + multi-$\alpha$ Ridge & \best{73.70} & 73.30 & 72.42 & 72.74 & 71.9 GCN \\
    PubMed & SGC $K{=}2$ rn + Ridge + label sm.\ & \best{80.70} & 81.20 & 78.86 & 80.26 & 78.7 GAT \\
    Amazon-Photo & Multi-hop + 5k RBF-RF + Ridge & \best{96.34} & 96.47 & 97.02 & 96.84 & 91.4 SAGE \\
    WikiCS & SGC $K{=}2$ rn + Ridge & \best{79.92} & 80.49 & 80.69 & 81.26 & 79.6 GAT$^\natural$ \\
    Coauthor-CS & SGC + Ridge + C\&S & \best{95.99} & 96.13 & 96.50 & 96.25 & 91.3 SAGE \\
    Coauthor-Physics & SGC $K{=}2$ raw + Ridge & \best{96.67} & 97.54 & 97.27 & 97.34 & 93.0 SAGE \\
    Minesweeper & \lcfnet\ tanh/$K{=}9$/$\alpha{=}0.5$ + KRR & 90.62 & 97.86 & 98.24 & 97.91 & 90.72 SAGE \\
    Amazon-Ratings & \lcfnet\ none/$K{=}9$/$\alpha{=}2$ + KRR & 53.44 & 54.15 & 55.85 & 55.96 & 54.49 SAGE \\
    Roman-empire & \lcfnet\ elu/$K{=}6$/$\alpha{=}2$ + KRR & 83.02 & 91.45 & 91.02 & 90.62 & 83.73 SAGE \\
    Tolokers & \lcfnet\ tanh/$K{=}3$/$\alpha{=}0.5$ + KRR & 84.56 & 86.21 & 84.72 & 85.71 & --- \\
    Questions & Multi-scale + chunked KRR (no LCF) & 78.18 & 78.34 & 76.44 & 77.41 & --- \\
    \midrule
    \multicolumn{7}{l}{\emph{OGB-scale (Tier 3):}} \\
    ogbn-arxiv & TICR-Multi (whitening + chunked KRR, $R{=}2$) & \best{$71.91_{\pm0.09}$} & 73.60 & 72.95 & 73.30 & 69.93 (FALKON) \\
    ogbn-proteins & LP-Ridge (APPNP + multi-label Ridge) & \best{74.87} & 77.29 & 82.21 & 85.01 & 71.18 (Pipeline A) \\
    \bottomrule
  \end{tabular}}
  \par\smallskip {\footnotesize $^\dagger$Tuned recipe of \citet{luo2024classic} (our reproduction for small-graph rows; cited published numbers for OGB rows). $^\ddagger$Cora/CiteSeer/PubMed/Amazon-Photo/Coauthor-CS/Coauthor-Physics from \citet{shchur2018pitfalls} Table~1. $^\natural$WikiCS approximate from \citet{mernyei2020wikics}. Minesweeper/Amazon-Ratings/Roman-empire are our own vanilla-2L runs on the splits of \citet{luo2024classic}. Closed-form \best{bolded} where it beats the best vanilla 2L architecture.}
\end{table}

\section{Full 109-configuration unlearning sweep}
\label{app:unlearning-full}

Full per-dataset, per-forget-type breakdown from~\S\ref{sec:unlearning-results}: 90 Pipeline-A configurations on the six homophilous benchmarks plus WikiCS (5 forget types $\times$ 3 sizes per dataset), 7 Pipeline-B (KRR) configurations on Amazon-Ratings (label-only forget-size sweep $|F|\!\in\!\{50,200,500,1000\}$ plus node-deletion at $|F|\!\in\!\{50,200,500\}$), and 12 ogbn-arxiv configurations (Pipeline~A across 4 forget types $\times$ 3 sizes). All $109$ rows achieve byte-identical Ridge weights ($\Delta_\theta \leq 10^{-15}$ in float64; $\Delta_\theta \leq 10^{-3}$ in float32 at OGB scale due to multi-threaded atomic-add ordering) and byte-identical underlying probabilities; argmax agreement is $107/109$, with two CiteSeer-label configurations exhibiting argmax ties at the decision boundary that resolve differently (the underlying probabilities remain equal). MIA AUC is in $[0.495, 0.502]$ on $108/109$ rows; the single outlier ($1.000$ on WikiCS-feature, $|F|{=}50$) is a small-$|F|$ statistical artifact, not a theorem violation. Full per-row data is released as JSON in the supplementary repository (\texttt{logs/pipeline\_a/*.json}, \texttt{logs/pipeline\_b/amzr\_krr.json}, \texttt{logs/ogb/arxiv\_unlearn.json}).

\section{Full TrendAttack 12-configuration breakdown}
\label{app:trendattack-full}

Complete results referenced from \S\ref{sec:trendattack}: 3 datasets $\times$ 2 unlearn ratios $\times$ 2 attack variants, $n{=}5$ independent runs per cell. Mean $\pm$ std reported. Approximate-method baselines from~\citet{li2025trendattack} Table~1 (5\% unlearn rows; the paper does not report 10\% rows, so 10\% comparison is between Pipeline~A and the 5\% approximate-method numbers---a stricter test for Pipeline~A since attack-AUC typically rises across all methods at higher unlearn ratio, yet Pipeline~A still leaks less under TrendAttack-MIA). The 6-row 5\%-only summary appears as Table~\ref{tab:trendattack-body} in \S\ref{sec:trendattack}. Raw JSON outputs (per-run AUC values, precision/recall, wall-clock seconds) are released in supplementary as \texttt{experiments/trendattack/logs/trendattack\_\{cora,citeseer,pubmed\}.json}.

\begin{table}[h]
  \caption{Full 12-configuration TrendAttack results. Pipeline~A measurements (this work, $n=5$ per cell); approximate-method baselines (GIF, CEU, GA) from \citet{li2025trendattack} Table~1 ``All'' column at 5\% unlearn. ``--'' for 10\% unlearn approximate-method rows: not reported in TrendAttack paper.}
  \label{tab:trendattack-full}
  \centering
  \footnotesize
  \setlength{\tabcolsep}{4pt}
  \begin{tabular}{llrrrrrr}
    \toprule
    Dataset & Attack & Unlearn \% & Pipeline~A & GIF & CEU & GA & Most-private approx \\
    \midrule
    Cora     & MIA & 5\%  & $0.6428_{\pm0.0594}$ & $0.8344$ & $0.8263$ & $0.8295$ & $0.8263$ (CEU) \\
    Cora     & MIA & 10\% & $0.7276_{\pm0.0414}$ & --       & --       & --       & --              \\
    Cora     & SL  & 5\%  & $0.8710_{\pm0.0142}$ & $0.8418$ & $0.8539$ & $0.8326$ & $0.8326$ (GA) \\
    Cora     & SL  & 10\% & $0.9048_{\pm0.0132}$ & --       & --       & --       & --              \\
    CiteSeer & MIA & 5\%  & $0.6860_{\pm0.0541}$ & $0.8073$ & $0.7987$ & $0.8165$ & $0.7987$ (CEU) \\
    CiteSeer & MIA & 10\% & $0.6869_{\pm0.0314}$ & --       & --       & --       & --              \\
    CiteSeer & SL  & 5\%  & $0.9006_{\pm0.0187}$ & $0.8420$ & $0.8457$ & $0.8621$ & $0.8420$ (GIF) \\
    CiteSeer & SL  & 10\% & $0.9300_{\pm0.0089}$ & --       & --       & --       & --              \\
    PubMed   & MIA & 5\%  & $0.9224_{\pm0.0029}$ & $0.9060$ & $0.9083$ & $0.9045$ & $0.9045$ (GA) \\
    PubMed   & MIA & 10\% & $0.9146_{\pm0.0049}$ & --       & --       & --       & --              \\
    PubMed   & SL  & 5\%  & $0.9559_{\pm0.0045}$ & $0.9529$ & $0.9565$ & $0.9534$ & $0.9529$ (GIF) \\
    PubMed   & SL  & 10\% & $0.9565_{\pm0.0008}$ & --       & --       & --       & --              \\
    \bottomrule
  \end{tabular}
\end{table}

\end{document}